\documentclass[10pt,twocolumn,letterpaper]{article}

\usepackage[pagenumbers]{cvpr}
\usepackage{booktabs,tabularx,array}
\definecolor{cvprblue}{rgb}{0.21,0.49,0.74}
\usepackage[pagebackref,breaklinks,colorlinks,allcolors=cvprblue]{hyperref}
\usepackage{multirow}
\usepackage{algorithm}
\usepackage[most]{tcolorbox}
\usepackage{algpseudocode}

\usepackage{amsthm}
\newtheorem{proposition}{Proposition}
\newtheorem{lemma}{Lemma}

\def\confName{CVPR}
\def\confYear{2027}

\title{Vision Transformers Need Separated Local and Global Information Processing}

\author{
Mingze Ma \qquad Ravi Garg \qquad Simon Lucey\\
Australian Institute for Machine Learning\\
Adelaide University\\
{\tt\small
\{mingze.ma, ravi.garg, simon.lucey\}@adelaide.edu.au}
}

\begin{document}
\maketitle
\begin{abstract}
Self-supervised vision transformers have become common foundation models for several applications. However, when trained at scale, these models suffer from degraded local feature representations, in which often uninformative image-background patches store global class information. While several interventions have been proposed to detect and alleviate this problem, the underlying mechanism responsible for the formation of these global tokens remains underexplored. In this work, we investigate the root cause of this information leakage from the global [CLS] token to local patch tokens. Our analysis identifies local patch tokens’ direct readout of global [CLS] information through self-attention as one of the direct causes of this phenomenon. We show that direct [CLS] readout precedes the emergence of high-norm tokens and attention sinks. Preventing the readout can remarkably enhance the local feature quality of vision transformers. Simply removing the problematic pathway from corrupted pretrained models improves performance on several dense tasks, including object discovery and zero-shot segmentation. We show that it is possible to further clean corrupted pretrained models through (i) value-projection-matrix surgery using calibration images or (ii) fine-tuning vision transformers (ViTs) to minimize the long tail of patch-to-[CLS] attention without using registers. Our experiments show that restricting patch-to-[CLS] attention consistently repairs ViTs exhibiting global-information-sink behavior and improves their performance on dense downstream tasks.
\end{abstract}    
\section{Introduction}
\begin{figure*}[!ht]
    \centering
    \includegraphics[width=0.9\textwidth]{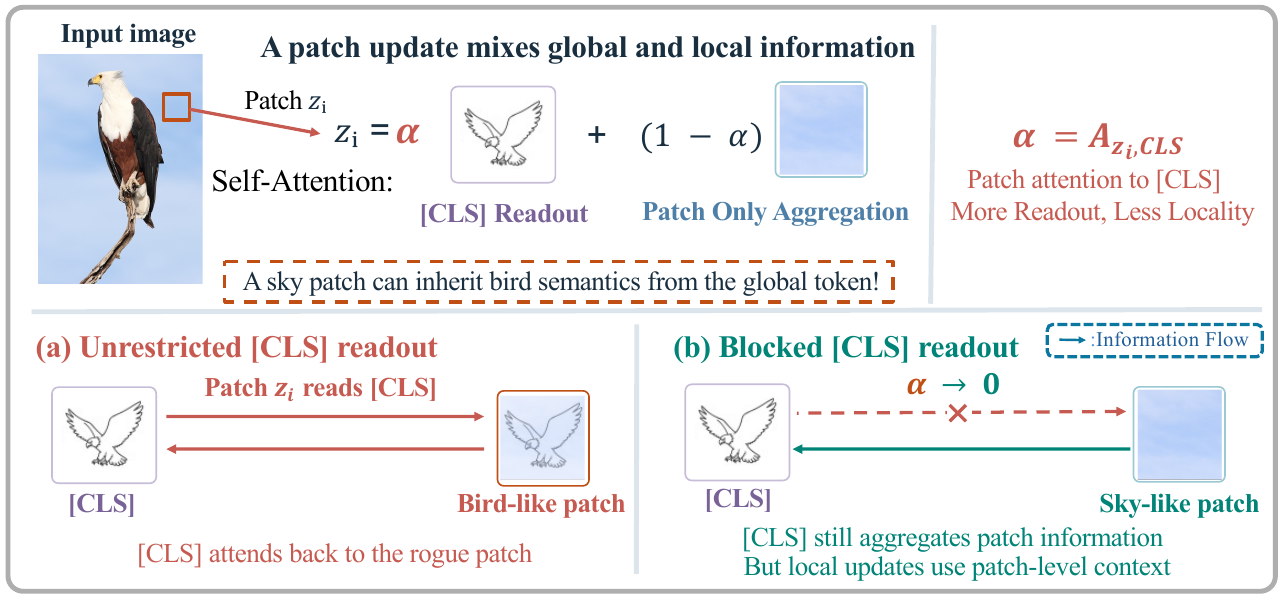}
\caption{\textbf{Excessive patch-to-[CLS] attention compromises local patch semantics.}
\textbf{Top:} A patch update combines global information from the [CLS] token with local contextual information from other patches. As patch-to-[CLS] attention ($a_{i,\mathrm{CLS}}$) increases, global information increasingly dominates the update, potentially causing a background patch to acquire foreground semantics.
\textbf{Bottom (a):} Unrestricted [CLS] readout allows local patches to become [CLS]-like, potentially attracting disproportionate [CLS]-to-patch attention ($a_{\mathrm{CLS},i}$) and creating a self-reinforcing cycle that leads to rogue tokens.
\textbf{Bottom (b):} Blocking direct [CLS] readout ($a_{i,\mathrm{CLS}}=0$) preserves local patch semantics while allowing the [CLS] token to continue aggregating information from patches.}
    \label{fig:overview}
\end{figure*}
Pretrained vision transformers (ViTs) \citep{Dosovitskiy2020AnII,vaswani2017attention,touvron2022deit,liu2021swin} produce powerful visual representations, yet their patch tokens can develop a striking pathology. Patches on nearly uniform or uninformative regions of an image can carry global class information instead of local semantic information. These \emph{rogue tokens} often compromise the spatial representations needed for dense downstream tasks and hamper the explainability of ViTs.

Several remedies have been proposed to mitigate this pathological behavior. Observing that background patches can mainly store global image information, Darcet et al. \citep{darcet2024vision} introduced register tokens to reroute this information away from image background patches. Trained registers have since become a gold-standard training-time remedy for such \emph{global-to-local information leakage}---though they may not be sufficient at larger scales. DINOv3~\cite{simeoni2025dinov3}, for instance, suggests that registers-based model may require additional regularization at larger training scales. Test-time registers~\cite{jiang2026vision} identify sparse-neuron outliers and relocate their activations into additional register tokens, demonstrating that such rerouting can be remarkably effective even after pretraining. SINDER~\cite{wang2024sinder} identifies dominant singular directions associated with defects in pretrained representations and corrects them through constrained fine-tuning. LazyStrike~\cite{shi2026vision} instead modifies [CLS] aggregation to reduce the influence of uninformative patches.
\emph{These interventions are largely motivated by the downstream manifestations of global-to-local information leakage, seeking to identify where the leaked information resides and mitigate its effects.} In this work, we shift the focus upstream and ask a more direct question:

\begin{tcolorbox}[
  colback=green!5!white,
  colframe=green!40!white,
  boxrule=0.45pt,
  arc=1.5pt,
  left=6pt,right=6pt,top=5pt,bottom=5pt,
  before skip=6pt,after skip=8pt
]
\noindent \textbf{What allows global information to enter local patches in the first place?}

\emph{We identify a direct mechanism contributing to rogue token formation in ViTs: repeated and unrestricted readout of information from the [CLS] token by local patch tokens.} This process can progressively contaminate local patch representations and turn the affected patch tokens to be [CLS]-like.
\end{tcolorbox}

Figure~\ref{fig:overview} illustrates how this contamination can develop and potentially reinforce itself. In standard ViTs, self-attention updates patch representations through attention-weighted aggregation of value vectors from patch tokens and the [CLS] token. Stronger [CLS] readout can therefore introduce more global information into local patch representations, potentially diluting their local semantics. Repeated across layers, this process can progressively make patch tokens more [CLS]-like. With finite attention logits, softmax assigns a strictly positive weight to an unmasked [CLS] token, keeping this readout pathway open. These increasingly [CLS]-like patch tokens may receive more attention from the [CLS] token if their query--key scores increase relative to those of other patches. Softmax can then concentrate attention on a small subset of these rogue tokens at the expense of other image regions. Global information written into local patch representations can subsequently be aggregated back into the [CLS] representation, creating a feedback loop that may further reinforce the contamination.


\begin{tcolorbox}[
  colback=green!5!white,
  colframe=green!40!white,
  boxrule=0.45pt,
  arc=1.5pt,
  left=6pt,right=6pt,top=5pt,bottom=5pt,
  before skip=6pt,after skip=8pt
]
\emph{\textbf{Central Thesis:} We argue that unrestricted patch readout of [CLS] is a harmful information-flow pathway and should be constrained during training of ViTs.}
\end{tcolorbox}

\noindent We leverage this central insight to repair pretrained ViTs with progressively greater flexibility to modify their learned representations and parameters (Table~\ref{tab:repair_levels}). All three interventions maintain the same principle of restricting direct [CLS] readout during downstream inference, while preserving [CLS] aggregation. \emph{Masking} prevents direct [CLS] readout at inference time without changing any model weights. \emph{Calibrated surgery} uses a small calibration set to correct recurring [CLS]-associated directions in value projections for patch tokens. \emph{Fine-tuning} allows broader adaptation of the pretrained network while preserving global semantics and regularizing excessive [CLS] readout.

\begin{figure*}[!ht]
    \centering
    \includegraphics[width=1\textwidth]{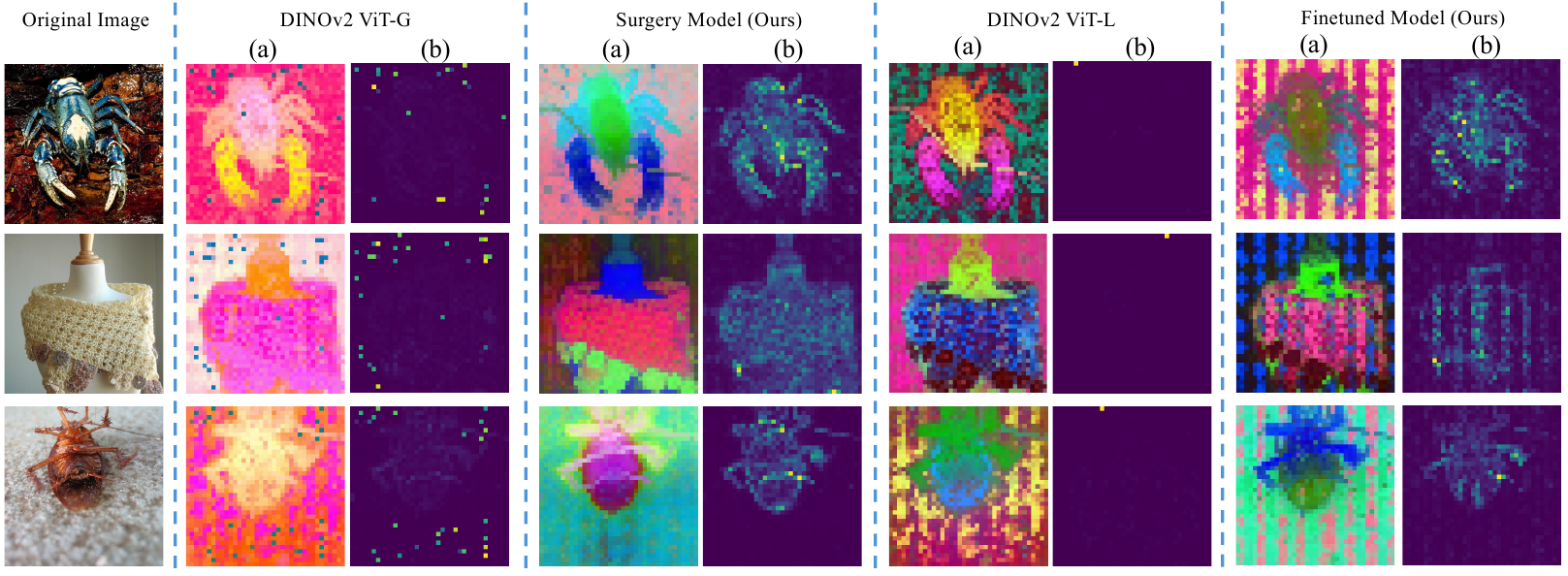}
    \caption{ We visualize (a) PCA of output patch representations and (b) CLS-to-patch attention maps. Our surgery method is applied to DINOv2 ViT-G and fine-tuning is applied to DINOv2 ViT-L. Both methods produce more spatially coherent patch representations and more informative CLS attention maps compared to their pretrained counterparts.}
    \label{fig:visual_over}
\end{figure*}

Progressively, more flexible interventions beyond just masking attention yield cleaner patch representations and further improvements on dense downstream tasks, as shown in \Cref{fig:visual_over}. Our experiments demonstrate how identifying the source of global-to-local information leakage allows us to correct pretrained ViTs increasingly effectively. 
\begin{tcolorbox}[
  colback=green!5!white,
  colframe=green!40!white,
  boxrule=0.45pt,
  arc=1.5pt,
  left=6pt,right=6pt,top=5pt,bottom=5pt,
  before skip=6pt,after skip=8pt
]
\emph{\textbf{We believe:} controlling the harmful information-flow pathway throughout large-scale pretraining could yield substantially stronger local representations without pathological rogues being formed from the outset.}
\end{tcolorbox}

 \noindent Further, we provide theoretical and empirical support for the hypothesis that excessive patch-to-[CLS] attention compromises local patch semantics. Under a simplified residual-attention model, we show that repeated [CLS] readouts progressively makes patch representations more [CLS]-like. We then observe the proposed chronology of rogue token formation in pretrained DINOv2-L across layers. \emph{Elevated [CLS] readout emerges first}, followed by sharp increases in norms of the rogue patches, which finally lead to disproportionate [CLS] attention to them. Crucially, we establish more than temporal ordering. Directly, restricting [CLS] readouts substantially suppresses both later abnormalities, demonstrating its causal contribution to rogue-patch formation empirically.

\begin{table}[!t]
\centering
\caption{Three levels of cleaning pretrained ViTs by restricting [CLS] readouts.}
\label{tab:repair_levels}
\footnotesize
\setlength{\tabcolsep}{3pt}
\renewcommand{\arraystretch}{1.2}
\begin{tabularx}{\columnwidth}{
@{}
>{\raggedright\arraybackslash}p{0.22\columnwidth}
>{\raggedright\arraybackslash}p{0.21\columnwidth}
>{\raggedright\arraybackslash}p{0.19\columnwidth}
>{\raggedright\arraybackslash}X
@{}
}
\toprule
\textbf{Method} &
\textbf{Correction scope} &
\textbf{Data} &
\textbf{Training signal} \\
\midrule
Masking &
None &
None &
None \\
\addlinespace[2pt]
Surgery &
Value projections &
500 images &
Calibration only \\
\addlinespace[2pt]
Fine-tune
(DINOv2) &
Entire ViT &
ImageNet-1K &
CLS distillation with penalty \\
\addlinespace[2pt]
Fine-tune
(OpenCLIP) &
Entire ViT &
LAION-2M &
Continued training with penalty \\
\bottomrule
\end{tabularx}
\end{table}

\noindent To summarize our central contributions are:
\begin{itemize}
  \item \emph{A novel upstream explanation of rogue token formation in ViTs.} We identify direct [CLS] readout as an early route from global information to local features, support its cumulative effect
  theoretically, and establish its causal contribution through intervention at different levels.
  \item \emph{A direct, training-free correction.} Restricting this pathway
  improves pretrained DINOv2 and OpenCLIP without additional tokens, data,
  or weight updates, including favorable comparisons against register-based
  methods on specified benchmarks.
  \item \emph{Weight repair from the same principle.} We extend the common inference-time restriction with calibration-guided value surgery and globally constrained fine-tuning to recover more useful local representations from pretrained models.
\end{itemize}

\section{Related Work}
Pretrained vision encoders \citep{assran2023self,assran2025v,he2022masked,balestriero2025lejepa,zhai2023sigmoid,radford2021learning,pmlr-v258-rusak25a,wang2024sclip,luo2023segclip, liu2023visual} provide general-purpose features through diverse self-supervised and language-supervised objectives.
DINO and DINOv2 learn representations through teacher--student self-distillation
\citep{caron2021emerging,oquab2023dinov2},
while CLIP learns a shared image--text embedding space through contrastive learning
\citep{radford2021learning} whose theoretical properties have also been studied
\citep{pmlr-v258-rusak25a}.
OpenCLIP enables large-scale contrastive pretraining using publicly available image--text data
\citep{cherti2023reproducible}.
These encoders have shown strong transferability across a range of vision tasks \citep{simeoni2021localizing,zhang2023tale,minderer2022simple,DBLP:conf/nips/ZuoWQS25,yang2024depth,wang2023tokencut,wang2024sclip, chefer2021transformer,liu2023visual}.

However, pretrained vision transformers \citep{oquab2023dinov2,touvron2022deit} are shown to develop pathological patch tokens with unusually large feature norms, attracting disproportionate attention from the [CLS] token.
\citet{darcet2024vision} observe that such tokens often occur in uninformative image regions and carry global information at the expense of local details, and introduce dedicated register tokens to accommodate internal computations. The central idea behind register is to re-route the global information away from the patches by providing a global scratchpad.
\citet{lappe2026register} investigate how register and [CLS] tokens contribute to the decoupling of global and local features.
DINOv3 introduces Gram anchoring to mitigate dense-feature degradation during prolonged large-scale pretraining
\citep{simeoni2025dinov3}.
More recently, \citet{marouani2026revisiting} specialize normalization layers and early query--key--value projections for [CLS] and patch tokens to improve dense representations. 

To remove rogue tokens from pretrained ViTs, SINDER \citep{wang2024sinder} identifies singular directions associated with feature defects and repairs them through fine-tuning.
\citet{yang2024denoising} propose denoising spatial feature maps, while \citet{chen2025vision} employ self-distillation to introduce registers into pretrained models.
Test-time registers \citep{jiang2026vision} redirect activations from identified MLP neurons to additional tokens without retraining.
LazyStrike \citep{shi2026vision} modifies global aggregation by selectively integrating patch information into the [CLS] token. Our work complements these approaches by studying direct [CLS] readouts during patch updates and using this information-flow pathway to improve local representations in pretrained ViTs.

A closely related architectural design to our idea of restricting global and local token mixing is used in CaiT \citep{touvron2021going}. CaiT, however were trained for supervised classification; and were never explicitly investigated this separation in relation to rogue-token formation. CaiT advocates for specialized layers with different weights to process CLS than those of patches as well. Our work investigates the consequences of unrestricted [CLS] readout upstream, repairs pretrained self-supervised ViTs, and retains shared attention projections without introducing separate processing stages.

Related work also studies the interpretability of CLIP representations
\citep{gandelsman2024interpreting}
and adapts CLIP to dense prediction through changes in patch aggregation and self-attention
\citep{luo2023segclip,wang2024sclip}.
ClearCLIP \citep{lan2024clearclip} modifies the final transformer block, while NACLIP \citep{hajimiri2025pay} encourages spatial locality in self-attention.
CLIPSelf \citep{wu2024clipself} transfers image-level recognition ability to local regions through self-distillation.

\section{Theoretical Intuition for Rogue Token}
\label{sec:cls_theory}

To understand how direct [CLS] readout can make local patch representations increasingly global, we consider a simplified transformer containing
only self-attention and residual connections. Let $h_i^\ell$ and
$h_C^\ell$ be the representations of patch $i$ and [CLS], respectively,
at layer $\ell$. We assume identity value transport and omit the MLP,
layer normalization, and learned projections. The patch update is
\begin{equation}
\label{eq:toy_patch_update}
h_i^{\ell+1}=h_i^\ell+a_{iC}^\ell h_C^\ell
+\sum_{j=1}^{N}a_{ij}^\ell h_j^\ell,
\end{equation}
where $a_{iC}^\ell$ denotes attention from patch query $i$ to the
[CLS] key, $a_{ij}^\ell\geq 0$, and
$a_{iC}^\ell+\sum_{j=1}^{N}a_{ij}^\ell=1$.

We further consider a regime in which the direction of the [CLS]
representation is stable. For tractability, suppose its attention
branch returns the same [CLS] representations, so the residual update becomes
\begin{equation}
\label{eq:toy_cls_update}
h_C^{\ell+1}=2h_C^\ell.
\end{equation}
This is an idealized fixed-direction update, rather than a claim that
CLS ceases to change in a pretrained ViT.

\noindent\textbf{Proposition 1 ([CLS]-induced alignment).}
Suppose $h_C^{\ell_0}\neq 0$ and there is a constant
$\varepsilon\in(0,1]$ such that
$a_{iC}^\ell\geq\varepsilon$ for every patch $i$ and every
$\ell\geq\ell_0$. Define the relative patch--[CLS] deviation as
\begin{equation}
\label{eq:toy_relative_deviation}
\rho_\ell=\frac{\max_i\|h_i^\ell-h_C^\ell\|_2}
{\|h_C^\ell\|_2}.
\end{equation}
Then
\begin{equation}
\label{eq:toy_contraction}
\rho_{\ell+1}\leq
\left(1-\frac{\varepsilon}{2}\right)\rho_\ell.
\end{equation}
Iterating this bound gives
$\rho_\ell\leq(1-\varepsilon/2)^{\ell-\ell_0}\rho_{\ell_0}$.
Consequently, $\rho_\ell\to0$ and
$\cos(h_i^\ell,h_C^\ell)\to1$ for every patch $i$ as
$\ell\to\infty$. The proof is given in the appendix.

The proposition formalizes how repeated [CLS] readout can progressively reduce the distinction between patches and the [CLS] token. Although it does not by itself explain why particular patches become rogue or why their norms grow sharply, we further analyze empirically whether elevated [CLS] readout appears before these effects in pretrained ViTs and whether directly restricting it works.

\section{Tracing the Origins of Rogue Tokens}
Previous studies have investigated anomalous ViT patch representations through their high-norm signatures and disproportionate involvement in [CLS]-to-patch attention. Register-based methods aim to redirect global information and internal computations away from image background patches into dedicated register tokens \cite{darcet2024vision,jiang2026vision}, while \cite{shi2026vision} aims to correct global aggregation itself by limiting contributions only from foreground patches, which carry predominantly class information. We instead ask: \emph{what causes patches to acquire disproportionately global characteristics in the first place, and can we prevent rogue token formation at the source?} 

\begin{figure*}[!ht]
    \centering
    \includegraphics[trim = 0 0 0 0, clip,  width=\textwidth]{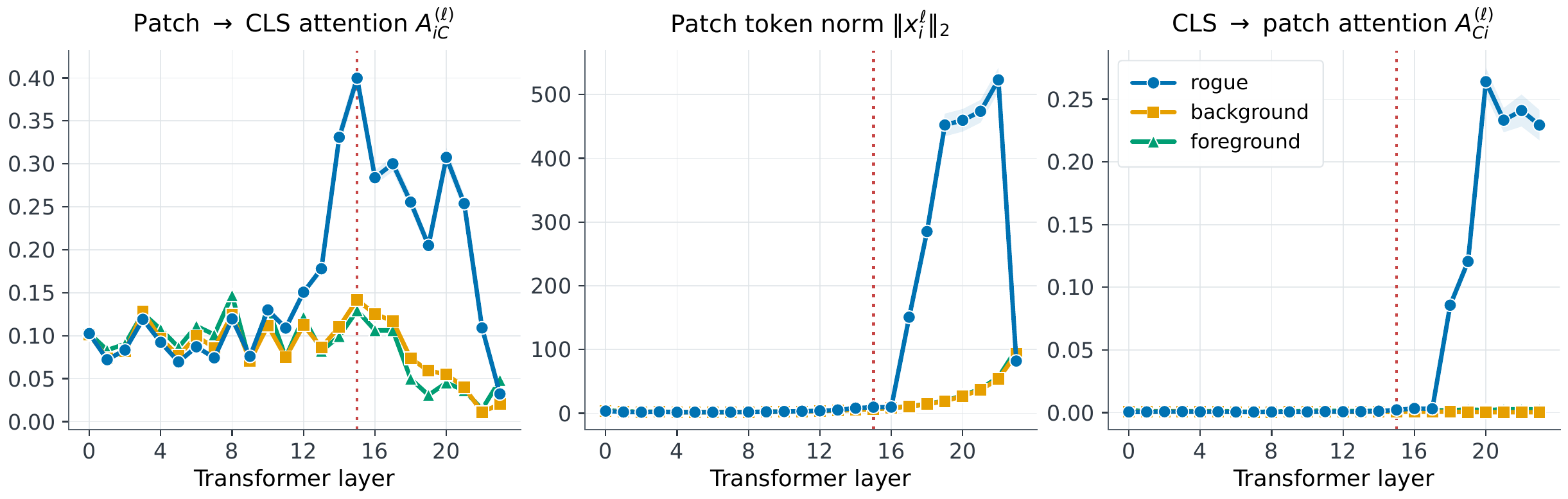}
    \caption{\textbf{Chronology of rogue token formation:} We track \textcolor{blue}{rogue}, \textcolor{orange}{background} and \textcolor{green}{foreground} image patches through all layers of DINOv2 ViT-L/14, and analyze patch $\rightarrow$ [CLS] attention (left), patch token norms (center) and [CLS] $\rightarrow$ patch attention (right). It can be observed that patches first read the [CLS] token disproportionately at layer 13, with the readout peaking at layer 15 (marked as red dotted line). Whereas the patch token norms and lazy aggregation of rogue tokens start in layer 18 -- i.e. much later than [CLS] token readout.}  
    \label{fig:outlier-attention-norm}
\end{figure*}
\begin{table}[!bh]
\centering
\caption{Controlled dose--response analysis of [CLS] readout in DINOv2-L/14. Token norms are measured after the penultimate transformer block, while [CLS]-to-patch attention is measured inside the final block. Rogue and ordinary background groups are fixed using the unmodified model.}
\label{tab:mechanistic_dose_response}
\setlength{\tabcolsep}{4pt}
\small
\begin{tabular}{|c|ccc|ccc|}
\hline
Leakage
& \multicolumn{3}{c|}{Average Token Norms}
& \multicolumn{3}{c|}{Average [CLS]-to-Patch Attn.} \\
\hline
 Cap
& Rogue & FG & BG
& Rogue & FG & BG \\
\hline
None
& 522.96 & 56.65 & 53.76
& 0.286 & 0.415 & 0.221 \\
\hline
0.3
& 354.27 & 55.30 & 52.58
& 0.247 & 0.410 & 0.258 \\
\hline
0.2
& 288.84 & 54.51 & 51.82
& 0.216 & 0.419 & 0.271 \\
\hline
0.1
& 173.09 & 52.84 & 50.28
& 0.139 & 0.450 & 0.293 \\
\hline
0
& 124.13 & 51.27 & 49.27
& 0.115 & 0.450 & 0.301 \\
\hline
\end{tabular}
\end{table}

To that end, we investigate direct feedback from the [CLS] token into local patch representations as a candidate mechanism, and causally test whether restricting this information flow alleviates both the emergence of high-norm representations and their subsequent disproportionate contribution to global aggregation.

\noindent \textbf{Experimental setup:} We analyze pretrained DINOv2-L/14 on 100 images from PASCAL VOC2007. Using ground-truth segmentation masks, patches containing at least 50\% non-void pixels are considered valid and are divided into three groups: foreground (FG), containing at least one foreground pixel; background includes rogue tokens (R), defined as pure-background patches whose final-layer attention from the [CLS] token, averaged across attention heads ($A_{Ci}$), exceeds the maximum corresponding attention weight among all foreground patches in the image; and ordinary background (BG), containing the remaining valid pure-background patches. These group assignments for patches are determined from the unmodified model and held fixed throughout the interventions. 

We track three quantities across the transformer depth for the aforementioned groups of patches: patch-to-[CLS] attention ($A_{iC}$), which controls direct readout of the global [CLS] representation by local patches; patch feature norms, a signature of outlier representations; and [CLS]-to-patch attention ($A_{Ci}$), which measures how [CLS] distributes its attention across image patch tokens and is commonly examined for transformer interpretability and object localization in ViTs \cite{caron2021emerging}. Note that while in the main manuscript we show DINOv2-L/14 analysis we observe broadly similar trends with DINOv2-G/14 and OpenCLIP-L/14 which are in the supplementary material. We answer the following three key questions with this analysis:


\begin{tcolorbox}[
    colback=blue!5,
    colframe=blue!35,
    boxrule=0.5pt,
    arc=1pt
]
\textbf{Q1: What is the first observed symptom of rogue token formation in a pretrained ViTs?}
\end{tcolorbox}

\noindent Figure \ref{fig:outlier-attention-norm} reveals that rogue patches first exhibit elevated patch-to-[CLS] attention $A_{iC}$ around blocks 13–15, reaching 0.40 attention at layer 15. At these depths, rogue token norms have not yet undergone pronounced amplification. Norms increase sharply around block 17 and keep increasing, while concentrated [CLS] aggregation $A_{Ci}$ emerges prominently around blocks 18–20 which is much later after multiple [CLS] readouts and patch aggregations. This indicates that [CLS] readouts precede the prominent high-norm phenotype in ViTs characterized by \cite{darcet2024vision} and further investigated through MLP activations by \cite{jiang2026vision}. It also suggests a possible precursor to the background-dominated patch aggregation studied by LazyStrike \cite{shi2026vision}. 

\textit{Our observations support the hypothesis that CLS readouts are the first detected causes of rogue-ness, successive readouts and MLPs lead to amplification of the rogue token norms, and multiple cycles of [CLS] readouts patch aggregation finally result in [CLS] disproportionately attending the rogue patches.} 

\begin{tcolorbox}[
    colback=blue!5,
    colframe=blue!35,
    boxrule=0.5pt,
    arc=1pt
]
\textbf{Q2: Is direct [CLS] readout causally responsible for norm amplification?}
\end{tcolorbox}
\noindent While the chronology of abnormalities aligns with the central hypothesis of our work, it does not directly imply causation. 
We further investigate the model's behavior via a controlled dose–response intervention. We progressively cap patch-to-[CLS] attention across all transformer blocks to restrict the [CLS] readouts. Then, we renormalize the attention assigned by each
patch token, without directly modifying those assigned by the [CLS] token. The modified attention weights are defined as follows:


\begin{equation}
\widetilde A_{ij} =
\begin{cases}
\min(A_{iC},\Omega),
& i\in\mathcal P,\ j=C,\\[6pt]
A_{ij}
\displaystyle\frac{1-\min(A_{iC},\Omega)}
{1-A_{iC}},
& i\in\mathcal P,\ j\neq C,\\[8pt]
\end{cases}
\label{eq:patch_cls_cap}
\end{equation}
Here, $A_{ij}$ and $\widetilde A_{ij}$ denote the original and modified attention weights from query patch token $i$ to key patch token $j$, respectively. $C$ denotes the [CLS] token, $\mathcal P$ represents the set of patch tokens, and $\Omega\in\{0.3,0.2,0.1,0\}$ specifies the maximum allowed patch-to-[CLS] attention. Attention exceeding $\Omega$ is proportionally redistributed among non-[CLS] tokens, preserving their relative attention weights and ensuring that each attention row sums to one. This redistribution preserves the relative attention weights among non-[CLS] tokens while ensuring that each attention row remains normalized, i.e., $\sum_j\widetilde A_{ij}=1$. Table \ref{tab:mechanistic_dose_response} shows that as we restrict the [CLS] readouts, mean rogue token norms decrease monotonically by 76.3\% under complete blocking. Foreground and ordinary-background norms decrease only moderately in comparison. We estimate the token norms in the penultimate transformer block, avoiding the sharp norm decrease in the final block for DINOv2 ViT-L/14 model.
 
It is important to note that unlike register-based approaches \citep{darcet2024vision,jiang2026vision}, which accommodate global information through additional tokens, our intervention suppresses the observed norm amplification without introducing additional tokens. The graded response provides causal evidence that direct [CLS] readout contributes substantially to the high-norm phenotype. \textit{Thus, elevated norm is not simply an artifact to detect and correct; it is a downstream outcome of a manipulable information-flow pathway.}

\begin{tcolorbox}[
    colback=blue!5,
    colframe=blue!35,
    boxrule=0.5pt,
    arc=1pt
]
\textbf{Q3: Can restricting global-to-local readout reduce subsequent lazy aggregation?}
\end{tcolorbox}
\noindent Remarkably, the control dose-response analysis in Table \ref{tab:mechanistic_dose_response} shows that the total [CLS] attention mass assigned to rogue patches decreases approximately 60\%. Meanwhile, both foreground attention mass ordinary-background mass increase. 
It should be noted that our intervention does not operate on the [CLS]-to-patch attention.
This is in stark contrast to \cite{shi2026vision} which directly intervene to selectively aggregate patch features into the global representation. Rather than changing how patch features are selected for aggregation, our analysis shows that \textit{restricting direct [CLS] readout into local patches does suppress the upstream formation of the rogue phenotype itself, thereby reducing its subsequent influence on global aggregation.}
\begin{tcolorbox}[
    colback=green!5,
    colframe=green!35,
    boxrule=0.5pt,
    arc=1pt,
    fontupper=\normalfont
]
\textbf{Mechanistic conclusion:} our analysis identifies elevated direct [CLS] readouts as causing rogue tokens' norm explosion, leading to these tokens participating disproportionately in [CLS] token formation. Restricting this pathway upstream causally suppresses both downstream phenomena, effectively increasing the fidelity of local representations.
\end{tcolorbox}

\section{Repairing Pretrained ViTs}
\label{sec:repair}
\begin{algorithm}[!ht]
  \caption{CLS-guided selection of surgery directions}
  \label{alg:direction-selection}
  \footnotesize
  \begin{algorithmic}[1]
  \Require Calibration images $\mathcal D=\{I_n\}_{n=1}^{N}$;
           surgery layers $\mathcal L_v$;
           thresholds $\gamma_{\rm img},\gamma_{\rm agg}$;
           per-head direction budget $K$
  \Ensure Selected direction matrices $\{D_{\ell h}\}$
  \State Define $\displaystyle
    \mathcal Z(a)_j=\frac{a_j-\operatorname{med}(a)}
       {\max\!\left(1.4826\operatorname{med}_k
          |a_k-\operatorname{med}(a)|,\delta\right)}$
  \ForAll{$\ell\in\mathcal L_v$, attention heads $h$}
    \State $W\gets W^{V}_{\ell h}\in\mathbb R^{d_h\times d}$;
           $w_j\gets W_{:,j}$ for $j=1,\ldots,d$
    \State $S,F,B\gets\mathbf 0\in\mathbb R^d$
    \For{$n=1,\ldots,N$}
      \State $x_n\gets$ CLS input to $W$ on image $I_n$
      \State $e_n\gets Wx_n$
      \State $s_j\gets\max(0,x_{n,j}w_j^\top e_n)$, $j=1,\ldots,d$
      \State $q_j\gets s_j/\max(\sum_{k=1}^d s_k,\delta)$
      \State $S\gets S+s$; $B\gets B+q$
      \State $F_j\gets F_j+\mathbf1\{\mathcal Z(s)_j\ge\gamma_{\rm img}\}$
    \EndFor
    \State $z\gets\mathcal Z(S)$
    \State $\pi\gets$ indices sorted by decreasing $F_j$,
           breaking ties by decreasing $B_j$
    \State $\mathcal J\gets[\,j\in\pi\mid z_j\ge\gamma_{\rm agg}\,]$
    \State $\mathcal J\gets$ first $\min(K,|\mathcal J|)$ indices of $\mathcal J$
    \State $\displaystyle D_{\ell h}\gets
       \left[\frac{w_j}{\max(\|w_j\|_2,1e-8)}\right]_{j\in\mathcal J}$
  \EndFor
  \State \Return $\{D_{\ell h}\}$
  \end{algorithmic}
\end{algorithm}
The intervention from our mechanistic analysis (Eq.\eqref{eq:patch_cls_cap}, $\Omega=0$) provides the common inference-time restriction we use in this paper. Keeping ViTs parameters unchanged, we call this our \emph{Masking} baseline. Note that pretraining with unrestricted [CLS] readouts demonstrates corrupt 
patch representations and embeds this contamination deeper into the model weights. Blocking the pathway afterward cannot directly undo this damage. We therefore investigate two additional refinements as discussed in \Cref{{tab:repair_levels}}.

\subsection{Calibration-Guided Value Surgery}
\label{sec:surgery}
We design a surgery method to exploit contaminated weights already learned by a pretrained model. Specifically, we identify recurring [CLS]-associated value column directions across randomly sampled images in ImageNet-1k and suppress their contributions to patch representations.
%
\paragraph{Selecting [CLS]-associated directions.}
We seek value-projection columns whose input-weighted contributions consistently align with the projected [CLS] feature across images. At each selected layer $\ell$ and attention head $h$, let $W=W^V_{\ell h}$ and denote its $j$-th column by $w_j$. For each of the 500 ImageNet-1K calibration images, we obtain the [CLS] input $x_n=[x_{n,j}]_j$ and its projected feature $e_n=Wx_n$. We assign each column the score $s_j=[x_{n,j}w_j^\top e_n]_+$, which measures the positive alignment of its weighted contribution $x_{n,j}w_j$ with $e_n$. We use robust Z-scores \citep{article} computed across columns to identify unusually large contributions. For each image, a column is counted as prominent if $\mathcal Z(s)_j\geq\gamma_{\rm img}$, and $F_j$ records the number of calibration images in which this condition holds. We also accumulate the raw scores in $S_j$ and the per-image normalized scores in $B_j$. These statistics capture complementary aspects of column importance: $F_j$ measures cross-image consistency but does not retain contribution magnitudes, whereas $S_j$ captures aggregate contribution strength. After processing all calibration images, we retain columns whose accumulated scores satisfy $\mathcal Z(S)_j\geq\gamma_{\rm agg}$. We rank the eligible columns by decreasing $F_j$, favoring contributions that are repeatedly prominent rather than large only on a few images. Ties are broken by decreasing $B_j$, which compares accumulated relative contributions after per-image normalization, reducing the influence of differences in overall score magnitude across images. We select up to $K$ columns, whose indices form $\mathcal J$. Finally, we construct $D_{\ell h}$ by stacking the selected normalized columns(Algorithm~\ref{alg:direction-selection}).
\paragraph{Positive-component value surgery.}
For each selected layer and head, let $D=D_{\ell h}$ contain the
individually normalized column directions returned by Algorithm~\ref{alg:direction-selection}.
We modify each patch value, but not the CLS value, as
\begin{equation}
  z_i=W^V_{\ell h}x_i+b^V_{\ell h},
  \qquad
  \widetilde z_i
  =z_i-\alpha D\left[D^\top z_i\right]_+,
\end{equation}
$[\cdot]_+$ denotes the element-wise positive-part operator. Only positive coefficients in $D^\top z_i$ are retained. Surgery acts only at layers $\mathcal L_v$. The CLS value is not modified directly, although subsequent CLS representations can change through the altered patches.

\subsection{Fine-Tuning for Broader Repair}
\label{sec:FT}

Unlike surgery, fine-tuning does not assume that contamination is concentrated in a small set of identifiable directions. Instead, it offers greater flexibility by allowing the model to adapt weights throughout the visual encoder. As shown in \Cref{fig:outlier-attention-norm}, rogue tokens exhibit abnormally large patch-to-CLS attention weights. Motivated by this observation, we propose a fine-tuning objective that can both preserves global semantic representations and penalizing heavy-tailed distributions of patch-to-CLS attention.
\begin{equation}
  \mathcal L_{\rm FT}=\mathcal L_{\rm global}
   +\lambda\mathcal L_{attention} .
  \label{eq:repair-ft}
\end{equation}
We use excess kurtosis as the measurement of the outliers of the patch-to-CLS attention distribution as follows:
\begin{equation}
\mathcal L_{attention}=\frac{1}{|\mathcal H|}
\sum_{\ell\in\mathcal H}
\left[
\frac{
\mathbb E_i\!\left[\left(A_{iC}^{\ell}-\mathbb E_j[A_{jC}^{\ell}]\right)^4\right]
}{
\left(\operatorname{Var}_i[A_{iC}^{\ell}]+\epsilon\right)^2
}
-3
\right]_+
\end{equation}
where $\mathcal H$ represents all selected heads. The regularizer penalizes positive excess kurtosis across patch-to-CLS attention, discouraging distributions dominated by unusually large deviations from the mean. Unlike directly penalizing large patch-to-CLS attention weights, this approach avoids overly aggressive suppression that could substantially disrupt the pretrained weights, particularly during the early stages of fine-tuning. For OpenCLIP, $\mathcal L_{\rm global}$ is the CLIP image--text contrastive objective with the text encoder frozen. For DINOv2-L, it is the L1 norm between the fine-tuned CLS feature and the CLS feature of a frozen pretrained DINOv2-L teacher.

\begin{table}[!b]
  \centering
  \caption{TokenCut localization performance using DINOv2
  variants. Each entry reports CorLoc ($\uparrow$) / mean best-GT
  bounding-box IoU (\%) ($\uparrow$) on VOC07, VOC12, and
  COCO20k. For each method and dataset, the two metrics are
  independently maximized over $(\tau,\epsilon)$.
  The best and second-best available results for each metric are marked in \textbf{bold} and
  \underline{underline}}
  \label{tab:dinov2_tokencut_combined}
  \footnotesize
  \setlength{\tabcolsep}{3.5pt}
  \renewcommand{\arraystretch}{1.08}
  \resizebox{\columnwidth}{!}{%
  \begin{tabular}{lccc}
    \toprule
    Method & VOC07 & VOC12 & COCO20k \\
    \midrule
    \multicolumn{4}{l}{\textbf{DINOv2 ViT-L/14}} \\
    DINOv2
    & 60.10 / 55.44
    & 63.38 / 57.64
    & 51.92 / 49.59 \\
    DINOv2 + Reg4
    & 65.61 / \underline{59.14}
    & \underline{69.12} / \textbf{62.12}
    & 54.57 / 51.28 \\
    \midrule
    Test-time register \cite{jiang2026vision}
    & 64.61 / 58.73
    & 67.85 / 61.21
    & \underline{55.08} / \textbf{52.04} \\
    Mask Only (Ours)
    & 61.66 / 56.52
    & 64.22 / 58.24
    & 54.70 / 51.28 \\
    Surgery (Ours)
    & \textbf{66.25} / \textbf{59.75}
    & 68.54 / \underline{61.72}
    & \textbf{55.16} / \underline{51.75} \\
    Finetune (Ours)
    & \underline{65.69} / 59.13
    & \textbf{69.12} / 61.56
    & 53.51 / 50.58 \\
    \midrule
    \midrule
    \multicolumn{4}{l}{\textbf{DINOv2 ViT-g/14}} \\
    DINOv2
    & 52.38 / 50.26
    & 61.93 / 56.97
    & 46.02 / 45.83 \\
    DINOv2 + Reg4
    & \underline{64.77} / \textbf{58.76}
    & \underline{67.58} / 61.24
    & 52.33 / 50.09 \\
    \midrule
    Test-time register \cite{jiang2026vision}
    & 62.22 / 57.55
    & 67.02 / \underline{61.27}
    & 51.60 / 49.85\\
    Mask Only (Ours)
    & 62.06 / 55.73
    & 64.77 / 57.40
    & \textbf{55.60} / \textbf{51.93} \\
    Surgery (Ours)
    & \textbf{65.33} / \underline{58.45}
    & \textbf{69.00} / \textbf{61.44}
    & \underline{55.28} / \underline{51.90} \\

    \bottomrule
  \end{tabular}%
  }
\end{table}

\begin{table}[!b]
  \centering
  \caption{Semantic segmentation on ImageNet-Segmentation with
  DINOv2 variants. All methods are evaluated on the full set of
  4,276 images at $518\times518$ resolution. Bold indicates the best result within each backbone.}
  \label{tab:imagenet_segmentation}
  \footnotesize
  \setlength{\tabcolsep}{3.5pt}
  \renewcommand{\arraystretch}{1}
  \resizebox{\columnwidth}{!}{%
  \begin{tabular}{lccc}
    \toprule
    Method & Pixel Acc. ($\uparrow$) & mIoU ($\uparrow$) & mAP ($\uparrow$)\\
    \midrule
    \multicolumn{4}{l}{\textbf{DINOv2 ViT-L/14}} \\
    DINOv2
    & 0.7830 & 0.5521 & 0.8535 \\
    DINOv2 + Reg4
    & 0.7268 & 0.5199 & 0.8135 \\
    \midrule
    Test-time register \cite{jiang2026vision}
    & 0.8205 & 0.6457 & \textbf{0.8599} \\
    Mask Only (Ours)
    & 0.7994 & 0.5928 & 0.8584 \\
    Surgery (Ours)
    & 0.8026 & 0.6126 & 0.8550 \\
    Finetune (Ours)
    & \textbf{0.8270} & \textbf{0.6482} & 0.8485 \\
    \midrule \midrule
    \multicolumn{4}{l}{\textbf{DINOv2 ViT-g/14}} \\ 
    DINOv2
    & 0.7251 & 0.4735 & 0.8394 \\
    DINOv2 + Reg4
    & 0.7988 & 0.6140 & 0.8516 \\
    \midrule
    Test-time register \cite{jiang2026vision}
    & 0.7423 & 0.5051 & 0.8410 \\
    Mask Only (Ours)
    & 0.8027 & 0.5990 & 0.8622 \\
    Surgery (Ours)
    & \textbf{0.8437} & \textbf{0.6829} & \textbf{0.8759} \\
    \bottomrule
  \end{tabular}%
  }
\end{table}

\begin{table*}[!ht]
  \centering
  \caption{Zero-shot semantic segmentation performance (mIoU, \% ($\uparrow$))
  using OpenCLIP ViT-L/14. We compare the five models: pretrained OpenCLIP, test-time register, applying mask for all layers of pretrained OpenCLIP, Surgery method and finetune method. 
  Background is excluded for VOC2012, COCO-Object, and Context59;
  void pixels are ignored. No segmentation head is trained.
  Bold and underline indicate the best and second-best
  results in each column, respectively.}
  \label{tab:vitl14_voc20_no_csa}
  \resizebox{\textwidth}{!}{%
  \begin{tabular}{lcccccc}
    \toprule
    Method
    & \shortstack{VOC2012 \citep{10.1007/s11263-014-0733-5}}
    & ADE20K \citep{Zhou_2017_CVPR} & COCO-Stuff10k \citep{Caesar_2018_CVPR} & COCO-Object \citep{lin2014microsoft} & Context59 \citep{Mottaghi_2014_CVPR} & Cityscapes \citep{cordts2016cityscapes} \\
    \midrule
    Baseline
    & 33.35 & 3.08 & 5.44 & 18.52 & 8.17 & 4.61 \\

    Test-time register \cite{jiang2026vision}
    & 35.13 & 3.11 & 5.45 & 18.64 & 8.23 & 4.58 \\

    Mask Only (Ours)
    & 40.51
    & \underline{3.68} & \underline{6.22}
    & 21.64 & \underline{9.45} & \underline{5.33} \\

    Surgery (Ours)
    & \underline{42.95}
    & 3.54 & 5.87
    & \underline{22.27} & 8.89 & 4.77 \\

    Finetune (Ours)
    & \textbf{45.11}
    & \textbf{4.92} & \textbf{7.48}
    & \textbf{24.51} & \textbf{11.56} & \textbf{7.48} \\
    \bottomrule
  \end{tabular}%
  }
\end{table*}

\section{Experiment}
In this section, we evaluate the proposed methods on downstream dense prediction tasks to verify that local information is preserved in the patch representations.
\subsection{Unsupervised Object Discovery with DINOv2}

We evaluate both surgery and fine-tune methods on unsupervised object discovery by using TokenCut method \citep{wang2022self} which is sensitive to the local token representations and representation coherence \citep{simeoni2025dinov3}. We evaluate the TokenCut on PASCAL 2007, 2012 \citep{Everingham2006ThePV} and COCO20k \citep{lin2014microsoft} . We sweep through the different $\tau \in \{0, 0.05, 0.10, \ldots, 0.60\}$ and $\epsilon=\{1e-4,1e-5\}$ for the output representations at the last layer.\Cref{tab:dinov2_tokencut_combined} reports TokenCut localization results on DINOv2 ViT-L/14 and ViT-g/14.
Both fine-tune and surgery improve localization performance over the original pretrained models across all three datasets, with surgery achieving competitive results compared to register-based methods. Fine-tuning also improves performance over the original ViT-L/14, achieving the highest CorLoc on VOC12. These results suggest that restricting patch-to-CLS attention can benefit object localization without introducing additional register tokens.

\subsection{Zero-shot Segmentation with DINOv2}
Unsupervised object discovery evaluates whether learned patch representations can identify semantically coherent object regions, with higher CorLoc indicating better object localization. To assess the interpretability of attention maps, which can be affected by rogue tokens, we use the mean CLS-to-patch attention from the last transformer layer. We then threshold the attention map at its mean value to obtain a binary foreground segmentation mask and evaluate its quality on ImageNet-Segmentation~\citep{chefer2021transformer}. As reported in \Cref{tab:imagenet_segmentation}, surgery on ViT-g/14 outperforms both trained and test-time registers across all metrics. On ViT-L/14, fine-tuning achieves higher Pixel Accuracy and mIoU than both register-based methods.


\subsection{OpenCLIP Zero-Shot Segmentation}
CLIP’s shared image–text embedding space enables zero-shot semantic segmentation by matching patch features to class text embeddings. We evaluate on six datasets including PASCAL VOC 2012\citep{10.1007/s11263-014-0733-5}, ADE20K \citep{Zhou_2017_CVPR}, COCO-Stuff10k \citep{Caesar_2018_CVPR}, COCO-Object \citep{lin2014microsoft}, Context59 \citep{Mottaghi_2014_CVPR} and Cityscapes \citep{cordts2016cityscapes} . We encode class names using an ensemble of text prompts and compute cosine similarities between the text embeddings and projected patch features. Each pixel is then assigned to the highest-scoring class. We report mean intersection-over-union (mIoU) as shown in \Cref{tab:vitl14_voc20_no_csa}. Our fine-tuned model achieves the best mIoU across all six datasets. Surgery also improves performance over baseline models on all datasets. Even masking alone provides consistent improvements over the original pretrained model. These results suggest that excessive patch-to-CLS attention allows global information to leak into local patches, and that simply restricting this information flow can substantially improve local representations.



\section{Conclusion and Limitations}
\paragraph{Conclusion.}
In this work, we investigate rogue-token formation in pretrained vision transformers and identify excessive patch-to-CLS attention as an important source of global information leakage into local patch representations. Mechanistic analysis and controlled interventions identify this pathway as an upstream cause, with high norms and degraded attention as downstream symptoms. We introduce three levels of intervention: mask, calibration-guided value surgery, and attention-regularized fine-tuning. Experiments show that these interventions improve local representations and dense prediction performance.
\paragraph{Limitations and Future Work.}
While we demonstrate that our methods enable rogue-token suppression and generalizes across pretrained vision transformers, our experiments focus on repairing pretrained models rather than training from scratch due to the substantial computational cost. we leave the investigation of restricting patch-to-CLS attention when train from scratch as future work. 



\newpage
{
    \small
    \bibliographystyle{ieeenat_fullname}
    \bibliography{main}
}
\newpage
\clearpage
\setcounter{page}{1}
\maketitlesupplementary
\section{Proof of Proposition 1}
\begin{proposition}[Patch-to-CLS collapse]
Let $h_{\mathrm{cls}}^\ell,h_1^\ell,\ldots,h_N^\ell\in\mathbb{R}^d$ denote one CLS token and $N$ patch tokens at layer $\ell\ge0$, with $h_{\mathrm{cls}}^0\neq0$. Suppose their representations evolve according to
\begin{align}
h_i^{\ell+1} &= h_i^\ell + a_{i,\mathrm{cls}}^\ell h_{\mathrm{cls}}^\ell + \sum_{j=1}^N a_{ij}^\ell h_j^\ell,\qquad i=1,\ldots,N,\\
h_{\mathrm{cls}}^{\ell+1} &= 2h_{\mathrm{cls}}^\ell,
\end{align}
where, for every $i$ and $\ell$,
\[
a_{i,\mathrm{cls}}^\ell\ge0,\qquad a_{ij}^\ell\ge0,\qquad a_{i,\mathrm{cls}}^\ell+\sum_{j=1}^N a_{ij}^\ell=1.
\]
Assume that there exists a constant $\epsilon\in(0,1]$ such that
\[
a_{i,\mathrm{cls}}^\ell\ge\epsilon,\qquad \forall i,\ell.
\]
Define the relative patch-to-CLS deviation by
\[
\rho_\ell:=\frac{\max_{1\le i\le N}\|h_i^\ell-h_{\mathrm{cls}}^\ell\|_2}{\|h_{\mathrm{cls}}^\ell\|_2}.
\]
Then
\[
\rho_\ell\le\left(1-\frac{\epsilon}{2}\right)^\ell\rho_0,
\]
and hence $\rho_\ell\to0$ as $\ell\to\infty$. Moreover, every patch representation is nonzero for all sufficiently large $\ell$ and satisfies
\[
\lim_{\ell\to\infty}\cos(h_i^\ell,h_{\mathrm{cls}}^\ell)=1,\qquad i=1,\ldots,N.
\]
\end{proposition}

To prove this proposition, we first present three lemmas (\Cref{lemma:1,lemma:2,lemma:3}) that establish the convergence properties required for the proof of Proposition 1.

\begin{lemma}
\label{lemma:1}
Define the patch-to-CLS deviation as
\[
e_i^\ell
:=
h_i^\ell-h_{\mathrm{cls}}^\ell.
\]
The deviation evolves according to
\[
e_i^{\ell+1}
=
e_i^\ell
+
\sum_{j=1}^{N}
a_{ij}^\ell e_j^\ell.
\]
\end{lemma}

\begin{proof}
Starting from the patch update,
\begin{align}
h_i^{\ell+1}
&=
h_i^\ell
+
a_{i,\mathrm{cls}}^\ell h_{\mathrm{cls}}^\ell
+
\sum_{j=1}^{N}
a_{ij}^\ell h_j^\ell.
\end{align}

Using
\[
h_j^\ell
=
h_{\mathrm{cls}}^\ell
+
e_j^\ell,
\]
we obtain
\begin{align}
h_i^{\ell+1}
&=
h_i^\ell
+
a_{i,\mathrm{cls}}^\ell h_{\mathrm{cls}}^\ell
+
\sum_{j=1}^{N}
a_{ij}^\ell
\left(
h_{\mathrm{cls}}^\ell+e_j^\ell
\right)
\\
&=
h_i^\ell
+
\left(
a_{i,\mathrm{cls}}^\ell
+
\sum_{j=1}^{N}a_{ij}^\ell
\right)
h_{\mathrm{cls}}^\ell
+
\sum_{j=1}^{N}
a_{ij}^\ell e_j^\ell.
\end{align}
Since
\[
a_{i,\mathrm{cls}}^\ell
+
\sum_{j=1}^{N}a_{ij}^\ell
=
1,
\]
we have
\[
h_i^{\ell+1}
=
h_i^\ell
+
h_{\mathrm{cls}}^\ell
+
\sum_{j=1}^{N}
a_{ij}^\ell e_j^\ell.
\]
Subtracting
\[
h_{\mathrm{cls}}^{\ell+1}
=
2h_{\mathrm{cls}}^\ell,
\]
gives
\begin{align}
e_i^{\ell+1}
&=
h_i^{\ell+1}
-
h_{\mathrm{cls}}^{\ell+1}
\\
&=
h_i^\ell
-
h_{\mathrm{cls}}^\ell
+
\sum_{j=1}^{N}
a_{ij}^\ell e_j^\ell
\\
&=
e_i^\ell
+
\sum_{j=1}^{N}
a_{ij}^\ell e_j^\ell.
\end{align}
\end{proof}

\begin{lemma}
\label{lemma:2}
Define
\[
E^\ell
:=
\max_{1\le i\le N}
\|e_i^\ell\|_2,
\]
and the relative patch-to-CLS deviation
\[
\rho_\ell
:=
\frac{E^\ell}
{\|h_{\mathrm{cls}}^\ell\|_2}.
\]
Suppose there exists a constant $\epsilon>0$ such that
\[
a_{i,\mathrm{cls}}^\ell
\ge
\epsilon,
\qquad
\forall i,\ell.
\]
Then
\[
\rho_{\ell+1}
\le
\left(
1-\frac{\epsilon}{2}
\right)
\rho_\ell
\]
and
\[
\frac{
\|h_i^\ell-h_{\mathrm{cls}}^\ell\|_2
}{
\|h_{\mathrm{cls}}^\ell\|_2
}
\longrightarrow
0.
\]
\end{lemma}

\begin{proof}
From \Cref{lemma:1} and the triangle inequality,
\begin{align}
\|e_i^{\ell+1}\|_2
&\le
\|e_i^\ell\|_2
+
\sum_{j=1}^{N}
a_{ij}^\ell
\|e_j^\ell\|_2
\\
&\le
E^\ell
+
\left(
\sum_{j=1}^{N}
a_{ij}^\ell
\right)
E^\ell.
\end{align}
Since
\[
\sum_{j=1}^{N}
a_{ij}^\ell
=
1-a_{i,\mathrm{cls}}^\ell,
\]
we obtain
\[
\|e_i^{\ell+1}\|_2
\le
\left(
2-a_{i,\mathrm{cls}}^\ell
\right)
E^\ell.
\]
Using
\[
a_{i,\mathrm{cls}}^\ell
\ge
\epsilon,
\]
we have
\[
E^{\ell+1}
\le
(2-\epsilon)E^\ell.
\]
On the other hand, $h_{\mathrm{cls}}^{\ell+1} = 2h_{\mathrm{cls}}^\ell,$ gives
\[
\|h_{\mathrm{cls}}^{\ell+1}\|_2
=
2\|h_{\mathrm{cls}}^\ell\|_2.
\]
Therefore,
\begin{align}
\rho_{\ell+1}
&=
\frac{E^{\ell+1}}
{\|h_{\mathrm{cls}}^{\ell+1}\|_2}
\\
&\le
\frac{(2-\epsilon)E^\ell}
{2\|h_{\mathrm{cls}}^\ell\|_2}
\\
&=
\left(
1-\frac{\epsilon}{2}
\right)
\rho_\ell.
\end{align}
Iterating this inequality yields
\[
\rho_\ell
\leq
\left(1-\frac{\epsilon}{2}\right)^\ell\rho_0.
\]
Since
\[0<1-\frac{\epsilon}{2}<1,\]
we have
\[
\rho_\ell\longrightarrow0\quad\text{as }\ell\to\infty.
\]
Therefore,
\[
\frac{
\|h_i^\ell-h_{\mathrm{cls}}^\ell\|_2
}{
\|h_{\mathrm{cls}}^\ell\|_2
}
\le \rho_\ell
\longrightarrow 0.
\]
\end{proof}

\begin{lemma}
\label{lemma:3}
Assume that $h_{\mathrm{cls}}^\ell \neq 0$ and that
\[
\frac{
\|h_i^\ell-h_{\mathrm{cls}}^\ell\|_2
}{
\|h_{\mathrm{cls}}^\ell\|_2
}
\longrightarrow 0.
\]
Then
\[
\frac{\|h_i^\ell\|_2}
{\|h_{\mathrm{cls}}^\ell\|_2}
\longrightarrow 1.
\]
\end{lemma}

\begin{proof}
By the reverse triangle inequality,
\[
\left|
\|h_i^\ell\|_2
-
\|h_{\mathrm{cls}}^\ell\|_2
\right|
\le
\|h_i^\ell-h_{\mathrm{cls}}^\ell\|_2.
\]
Dividing both sides by
$\|h_{\mathrm{cls}}^\ell\|_2>0$ gives
\[
\left|
\frac{\|h_i^\ell\|_2}
{\|h_{\mathrm{cls}}^\ell\|_2}
-1
\right|
\le
\frac{
\|h_i^\ell-h_{\mathrm{cls}}^\ell\|_2
}{
\|h_{\mathrm{cls}}^\ell\|_2
}.
\]
The right-hand side converges to zero (\Cref{lemma:2}). Therefore,
\[
\frac{\|h_i^\ell\|_2}
{\|h_{\mathrm{cls}}^\ell\|_2}
\longrightarrow 1.
\]
\end{proof}

\begin{proof}[\textbf{Proof for Proposition 1:}]
By \Cref{lemma:2}, we know
\[
\frac{
\|h_i^\ell-h_{\mathrm{cls}}^\ell\|_2
}{
\|h_{\mathrm{cls}}^\ell\|_2
}
\longrightarrow
0.
\]
Writing
\[
h_i^\ell
=
h_{\mathrm{cls}}^\ell
+
e_i^\ell,
\]
For every patch token $i$, we have
\[
\frac{
\|h_i^\ell-h_{\mathrm{cls}}^\ell\|_2
}{
\|h_{\mathrm{cls}}^\ell\|_2
}
\le \rho_\ell
\longrightarrow 0.
\]
Equivalently,
\[
\left\|
\frac{h_i^\ell}{\|h_{\mathrm{cls}}^\ell\|_2}
-
\frac{h_{\mathrm{cls}}^\ell}
{\|h_{\mathrm{cls}}^\ell\|_2}
\right\|_2
\longrightarrow 0.
\]
This shows that, after normalization by the CLS norm, the patch
representation approaches the normalized CLS representation. Notice,
however, that
\[
\frac{h_i^\ell}{\|h_{\mathrm{cls}}^\ell\|_2}
\]
is not necessarily a unit vector. We therefore need to relate it to the
unit-normalized patch representation.

By \Cref{lemma:3},
\[
\frac{\|h_i^\ell\|_2}
{\|h_{\mathrm{cls}}^\ell\|_2}
\longrightarrow 1.
\]
In particular, $h_i^\ell\neq 0$ for all sufficiently large $\ell$.
Moreover,
\begin{align}
&
\left\|
\frac{h_i^\ell}{\|h_i^\ell\|_2}
-
\frac{h_{\mathrm{cls}}^\ell}
{\|h_{\mathrm{cls}}^\ell\|_2}
\right\|_2
\\
&\le
\left\|
\frac{h_i^\ell}{\|h_i^\ell\|_2}
-
\frac{h_i^\ell}{\|h_{\mathrm{cls}}^\ell\|_2}
\right\|_2
+
\left\|
\frac{h_i^\ell}{\|h_{\mathrm{cls}}^\ell\|_2}
-
\frac{h_{\mathrm{cls}}^\ell}
{\|h_{\mathrm{cls}}^\ell\|_2}
\right\|_2
\\
&=
\left|
1-
\frac{\|h_i^\ell\|_2}
{\|h_{\mathrm{cls}}^\ell\|_2}
\right|
+
\frac{
\|h_i^\ell-h_{\mathrm{cls}}^\ell\|_2
}{
\|h_{\mathrm{cls}}^\ell\|_2
}.
\end{align}

The first term converges to zero by Lemma 3, while the second term is
bounded above by $\rho_\ell$ and therefore also converges to zero.
Consequently,
\[
\left\|
\frac{h_i^\ell}{\|h_i^\ell\|_2}
-
\frac{h_{\mathrm{cls}}^\ell}
{\|h_{\mathrm{cls}}^\ell\|_2}
\right\|_2
\longrightarrow 0.
\]
Since both vectors are unit vectors, we have
\begin{align}
\left\|
\frac{h_i^\ell}{\|h_i^\ell\|_2}
-\frac{h_{\mathrm{cls}}^\ell}{\|h_{\mathrm{cls}}^\ell\|_2}\right\|_2^2
&=2-2\frac{(h_i^\ell)^\top h_{\mathrm{cls}}^\ell}{\|h_i^\ell\|_2\|h_{\mathrm{cls}}^\ell\|_2}\\
&=2-2\cos
\left(h_i^\ell,h_{\mathrm{cls}}^\ell\right).
\end{align}
The left-hand side converges to zero, and hence
\[
\cos
\left(
h_i^\ell,h_{\mathrm{cls}}^\ell
\right)
\longrightarrow 1.
\]
\end{proof}

\section{Additional Mechanistic Analysis}
In this section, we extend our analysis to OpenCLIP and DINOv2 ViT-g to verify that our findings are not specific to a particular model size or training strategy.

\subsection{DINOv2 ViT-g/14}
\begin{figure*}[!h]
    \centering
    \includegraphics[trim = 0 0 0 0, clip,  width=\textwidth]{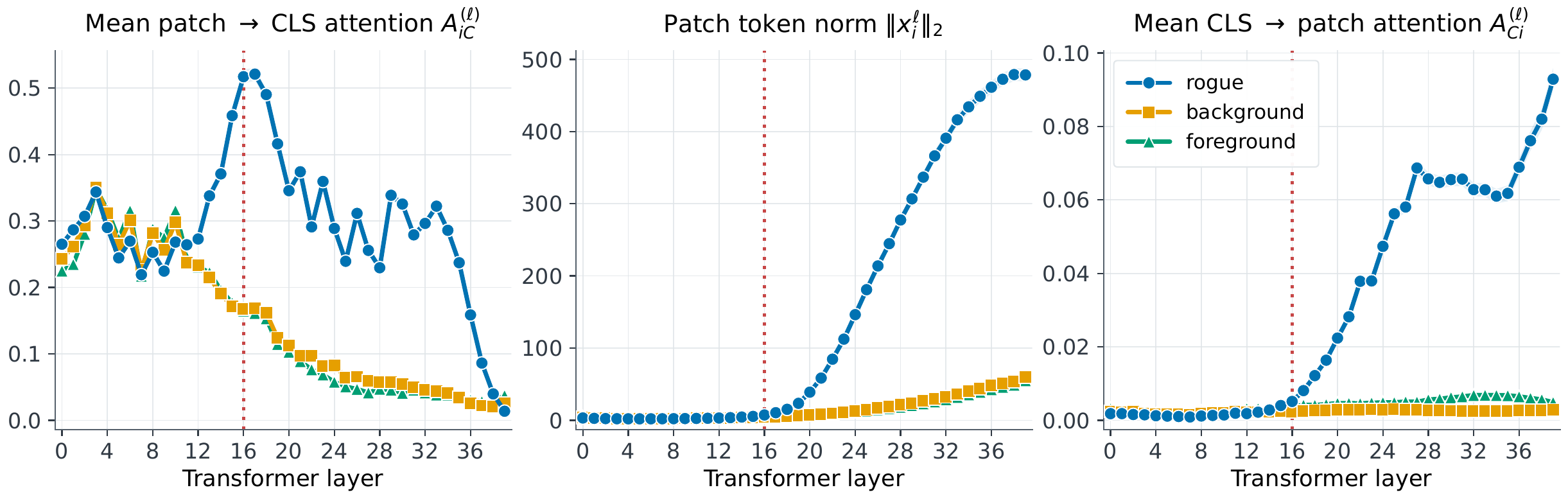}
    \caption{\textbf{Chronology of rogue token formation:} We track \textcolor{blue}{rogue}, \textcolor{orange}{background} and \textcolor{green}{foreground} image patches through all layers of DINOv2 ViT-G/14, and analyze patch $\rightarrow$ [CLS] attention (left), patch token norms (center) and [CLS] $\rightarrow$ patch attention (right) as a proxy for the model's explainability. It can be observed that patches first read the [CLS] token disproportionately in layer 12, and the readout peaks at layers 16--17 (marked as red dotted line) whereas the patch token norms and lazy aggregation of rogue tokens start in layer 17 onward -- i.e., much later than [CLS] token readout.}  
    \label{fig:vitgoutlier-attention-norm}
\end{figure*}

We conduct the same analysis on DINOv2 ViT-g/14. As shown in \Cref{fig:vitgoutlier-attention-norm}, the rogue token exhibits a substantial increase in patch-to-[CLS] attention starting around layer 12, reaching its peak near layers 16--17. In contrast, the rogue token norm increases sharply only in later layers, followed by an increase in [CLS]-to-patch attention. This ordering is consistent with our observations in DINOv2 ViT-L/14, further supporting the hypothesis that excessive patch-to-[CLS] attention precedes the formation of high-norm patch outliers.
\begin{table}[!ht]
\centering
\caption{Controlled dose--response analysis of [CLS] readout in
DINOv2-Giant.  All conditions use the same 500 VOC2007 images. Token norms and [CLS]-to-patch attention are measured inside block 39 (Layer index starts from 0). Foreground/background use patch centers inside/outside the union of VOC boxes, with the rogue excluded from ordinary background. All groups of patches are fixed for different caps. Attention probabilities are summed over group patches within each image, then averaged over images.}
\label{tab:mechanistic_dose_response_giant}
\setlength{\tabcolsep}{4pt}
\small
\begin{tabular}{|c|ccc|ccc|}
\hline
Leakage & \multicolumn{3}{c|}{Token Norms} & \multicolumn{3}{c|}{[CLS]-to-Patch Attn.} \\
Cap & Rogue & FG & BG & Rogue & FG & BG \\
\hline
None & 478.4 & 55.09 & 59.65 & 0.09288 & 0.4221 & 0.3907 \\
\hline
0.3 & 308.2 & 55.68 & 59.14 & 0.06797 & 0.4289 & 0.3825 \\
\hline
0.2 & 228.5 & 54.82 & 57.51 & 0.04746 & 0.4368 & 0.3825 \\
\hline
0.1 & 119.1 & 51.38 & 53.44 & 0.01459 & 0.4960 & 0.3977 \\
\hline
0 & 53.12 & 49.52 & 50.09 & 0.00353 & 0.4635 & 0.5107 \\
\hline
\end{tabular}
\end{table}
We further investigate this relationship through a controlled dose--response intervention, as shown in \Cref{tab:mechanistic_dose_response_giant}. As the maximum allowable patch-to-[CLS] attention is progressively reduced, the rogue token norm decreases substantially from 478.40 to 53.12 (88.9\%), while foreground and background patch norms exhibit comparatively smaller reductions of 10.1\% and 16.0\%, respectively. Meanwhile, [CLS]-to-rogue attention mass decreases sharply from 0.09288 to 0.00353 (96.2\%), accompanied by increased attention to foreground and ordinary background patches. This is consistent with the trend observed in DINOv2 ViT-L/14, further suggesting that the phenomenon persists across different model scales. 
\subsection{OpenCLIP ViT-L/14}
\begin{figure*}[!ht]
    \centering
    \includegraphics[trim = 0 0 0 0, clip,  width=\textwidth]{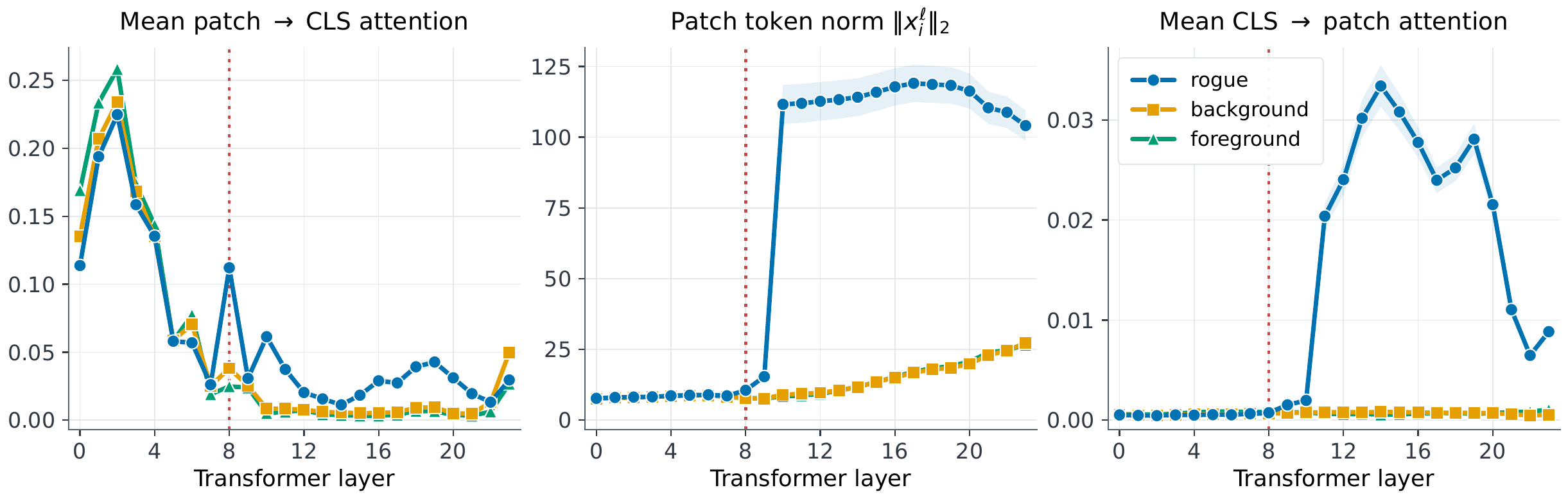}
    \caption{\textbf{Chronology of rogue token formation:} We track \textcolor{blue}{rogue}, \textcolor{orange}{background} and \textcolor{green}{foreground} image patches through all layers of OpenCLIP ViT-L/14, and analyze patch $\rightarrow$ [CLS] attention (left), patch token norms (center) and [CLS] $\rightarrow$ patch attention (right) as a proxy for the model's explainability. It can be observed that patches first read the [CLS] token disproportionately in layer 8, whereas the patch token norms and lazy aggregation of rogue tokens start in layer 10 onward -- i.e., much later than CLS token readout.}  
    \label{fig:ocoutlier-attention-norm}
\end{figure*}

As shown in \Cref{fig:ocoutlier-attention-norm}, although the pattern is less pronounced than in DINOv2 ViT-L/14, the rogue token exhibits elevated patch-to-[CLS] attention relative to other patches across layers 8--12 and 15--21. Notably, this attention begins to increase at layer 8,
preceding the sharp increase in rogue token norm around
layer 10. The increase in [CLS]-to-patch attention occurs even later. Despite differences in the magnitude of these effects, the ordering remains consistent with our observations in DINOv2 ViT-L/14.
\begin{table}[!ht]
\centering
\caption{Controlled dose--response analysis of [CLS] readout in OpenCLIP ViT-L/14. All conditions use the same 500 VOC2007 images. Token norms and [CLS]-to-patch attention are measured inside block 23 (Layer index starts from 0). Foreground/background use patch centers inside/outside the union of VOC boxes, with the rogue excluded from ordinary background. All groups of patches are fixed for different caps. Attention probabilities are summed over group patches within each image, then averaged over images. }
\label{tab:mechanistic_dose_response_openclip_vitl_train500_background_rogue_mass}
\setlength{\tabcolsep}{4pt}
\small
\begin{tabular}{|c|ccc|ccc|}
\hline
Leakage & \multicolumn{3}{c|}{Token Norms} & \multicolumn{3}{c|}{[CLS]-to-Patch Attn.} \\
Cap & Rogue & FG & BG & Rogue & FG & BG \\
\hline
None & 108.8 & 24.76 & 24.64 & 0.00885 & 0.5173 & 0.3745 \\
\hline
0.3 & 105.5 & 24.67 & 24.47 & 0.00825 & 0.5167 & 0.3731 \\
\hline
0.2 & 102.0 & 24.57 & 24.31 & 0.00825 & 0.5160 & 0.3715 \\
\hline
0.1 & 96.68 & 24.41 & 24.06 & 0.00814 & 0.5159 & 0.3683 \\
\hline
0 & 89.22 & 24.07 & 23.69 & 0.00793 & 0.5191 & 0.3628 \\
\hline
\end{tabular}
\end{table}

We also investigate this relationship through a controlled dose--response intervention, as shown in \Cref{tab:mechanistic_dose_response_openclip_vitl_train500_background_rogue_mass}. Consistent with our observations in DINOv2 ViT-L/14, progressively restricting patch-to-[CLS] attention reduces the rogue token norm from 108.80 to 89.22 (18.0\%), while foreground and background patch norms decrease by only 2.8\% and 3.9\%, respectively. Meanwhile, [CLS]-to-rogue attention mass also decreases from 0.00885 to 0.00793 (10.4\%), while [CLS]-to-foreground attention remains nearly unchanged and [CLS]-to-background attention decreases slightly. The trend observed in DINOv2 therefore persists in OpenCLIP, albeit with less pronounced changes, potentially reflecting differences in their pretraining strategies. These findings further support the hypothesis that excessive patch-to-[CLS] attention contributes to the formation of high-norm patch outliers and is associated with their subsequent attention from the [CLS] token.

\section{Detailed Implementation of Methods}
In this section, we provide the detailed parameters that we adopt for our methods.
\subsection{Surgery}
For the surgery method, we present the parameters in \Cref{tab:surgery_parameters}. We choose the surgery layers based on our mechanistic analysis where the patch-to-[CLS] attention is disproportionately large. 
\begin{table}[!h]
\centering
\scriptsize
\setlength{\tabcolsep}{3pt}
\renewcommand{\arraystretch}{1.1}
\begin{tabular}{@{}lccc@{}}
\toprule
& \shortstack{DINOv2\\ViT-L/14}
& \shortstack{DINOv2\\ViT-g/14}
& \shortstack{OpenCLIP\\ViT-L/14} \\
\midrule
$\mathcal L_v$
& $\{13,\ldots,16\}$
& $\{14,\ldots,23\}$
& $\{6,7,8\}$ \\
Mask layers
& 0--23 & 0--39 & 0--23 \\
$\alpha$
& 2.0 & 3.8 & 7.0 \\
$\gamma_{\rm img}$
& 70 & 60 & 30 \\
$\gamma_{\rm agg}$
& 70 & 60 & 30 \\
$K$
& 3 & 2 & 1 \\
Calib. resolution
& $224\times224$
& $518\times518$
& $224\times224$ \\
\bottomrule
\end{tabular}
\caption{
Hyperparameters for CLS-guided direction selection and surgery.
Surgery uses soft projection on value-projection input
columns and positive-only removal. Mask layers indicate where patch-to-CLS
attention is blocked. Layer indices start from 0.
Calibration resolutions are used for direction selection.
}
\label{tab:surgery_parameters}
\end{table}

\subsection{Fine-tune}
For the fine-tuning, we show the hyperparameters in \Cref{tab:finetune_parameters} for DINOv2 ViT-L and \Cref{tab:openclip_previous_finetune_parameters} for OpenCLIP ViT-L.
\begin{table}[!h]
    \centering
    \footnotesize
    \setlength{\tabcolsep}{4pt}
    \begin{tabular}{lc}
        \toprule
        Hyperparameter & DINOv2 ViT-L/14 \\
        \midrule
        \multicolumn{2}{l}{\textit{Optimization and training budget}} \\
        Optimizer & AdamW \\
        AdamW $(\beta_1,\beta_2)$ & $(0.9,0.999)$ \\
        AdamW $\epsilon$ & $10^{-8}$ \\
        learning rate & $10^{-6}$ \\
        Warmup & 50 steps \\
        Weight decay & 0.04 \\
        Gradient clipping (norm) & 1.0 \\
        Effective batch $B_{\rm eff}$ & 512 \\
        Optimizer updates $T$ & 1000 \\
        Training input & $224\times224$ \\
        \midrule
        \multicolumn{2}{l}{\textit{Losses and regularization}} \\
        CLS distillation loss & $\ell_1$ \\
        CLS loss weight $\lambda_{\rm cls}$ & 1 \\
        Target kurtosis weight $\lambda_{\rm kurt}$ & 1000 \\
        Kurtosis-penalty layers & 14--23 \\
        Patch-to-CLS mask layers & 0--23 \\
        \bottomrule
    \end{tabular}
    \caption{Fine-tuning hyperparameters for the DINOv2 ViT-L checkpoint.}
    \label{tab:finetune_parameters}
    
\end{table}

\begin{table}[!h]
    \centering
    \footnotesize
    \setlength{\tabcolsep}{4pt}
    \begin{tabular}{lc}
        \toprule
        Hyperparameter & OpenCLIP ViT-L/14 \\
        \midrule
        \multicolumn{2}{l}{\textit{Optimization and training budget}} \\
        Initialization & LAION-2B B82K \\
        Training data & LAION2M \\
        Optimizer & AdamW \\
        AdamW $(\beta_1,\beta_2)$ & $(0.9,0.98)$ \\
        AdamW $\epsilon$ & $10^{-6}$ \\
        Peak learning rate & $10^{-5}$ \\
        Learning-rate warmup & 50 steps \\
        Weight decay & 0.04 \\
        Gradient clipping (norm) & 1.0 \\
        Effective batch $B_{\rm eff}$ & 1024 \\
        Training epochs & 10 \\
        Optimizer updates $T$ & 19,540 \\
        Training input & $224\times224$ \\
        Text encoder & Frozen \\
        Image projection & Frozen \\
        \midrule
        \multicolumn{2}{l}{\textit{Losses and regularization}} \\
        Image--text loss & CLIP contrastive loss \\
        Contrastive loss weight $\lambda_{\rm CLIP}$ & 1 \\
        Kurtosis regularizer & Attention excess kurtosis \\
        Kurtosis weight $\lambda_{\rm kurt}$ & 1 \\
        Kurtosis-penalty layers & 9--14 \\
        Patch-to-CLS mask layers & 0-23 \\
        \bottomrule
    \end{tabular}
    \caption{Fine-tuning hyperparameters for the OpenCLIP ViT-L checkpoint.}
    \label{tab:openclip_previous_finetune_parameters}
\end{table}

\section{Additional Visualization}
In this section, we provide additional visualizations of [CLS]-to-patch attention maps and patch norms to compare the extent of rogue-token behavior across different methods.

\subsection{DINOv2 ViT-L/14}
\begin{figure}[!h]
    \centering
    \includegraphics[width=\columnwidth]{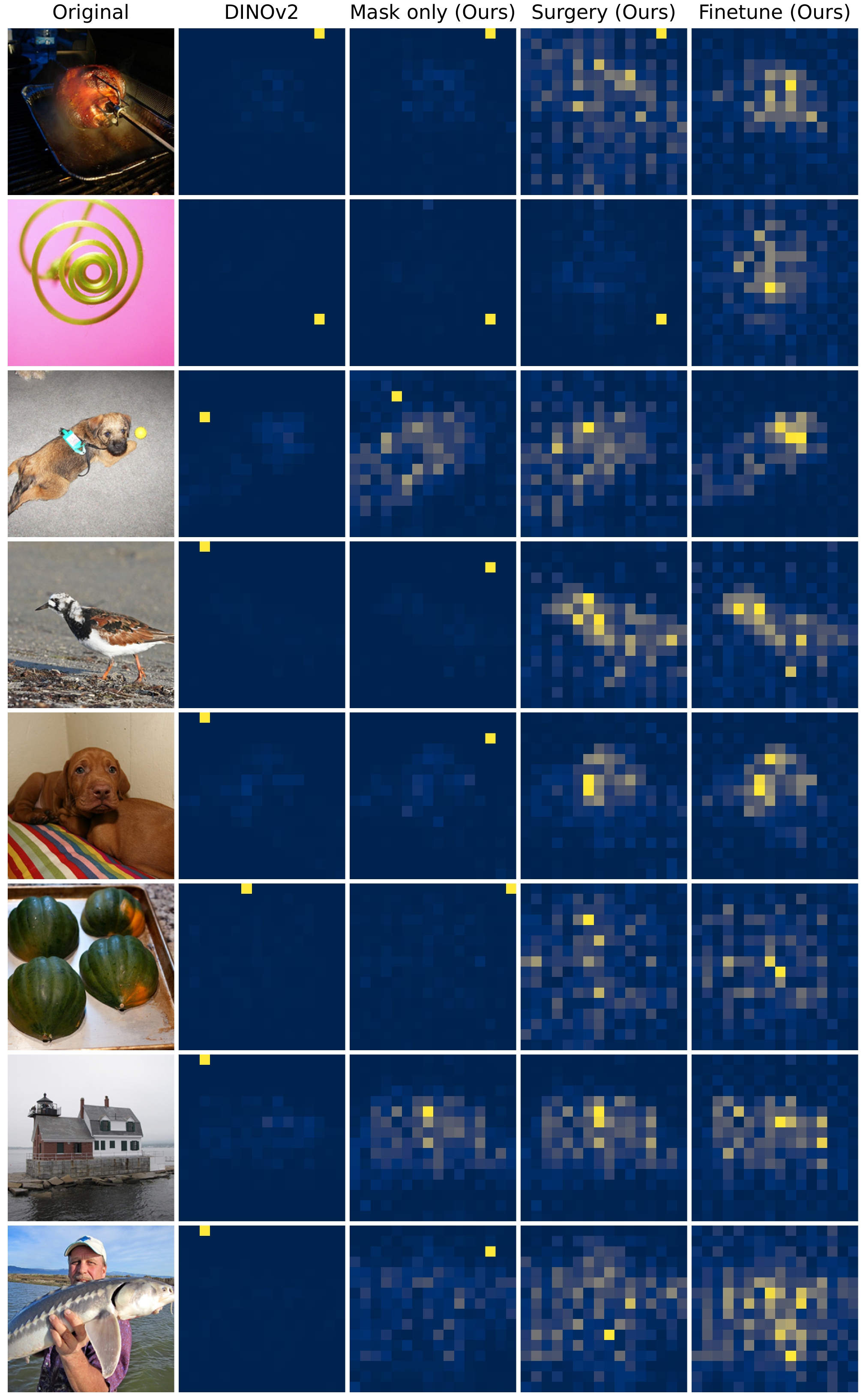}
    \caption{Comparison of [CLS]-to-patch attention across three proposed clean-up methods for DINOv2 ViT-L/14. Each heatmap is independently normalized using its own minimum and maximum values. Yellow indicates higher values, while blue indicates lower values.}
    \label{fig:vitlatt}
\end{figure}

\begin{figure}[!h]
    \centering
    \includegraphics[width=\columnwidth]{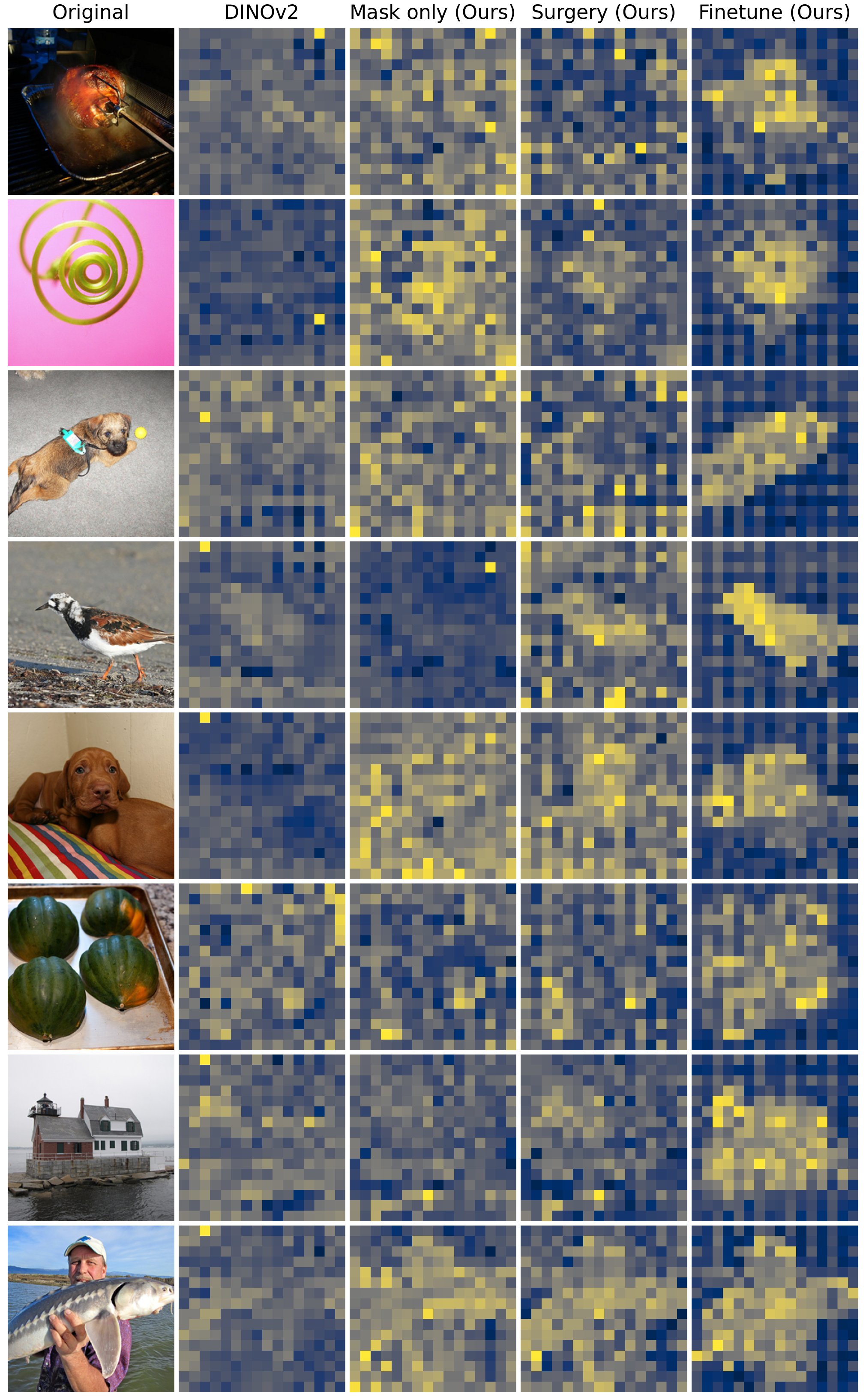}
    \caption{Comparison of patch norm across three proposed clean-up methods for DINOv2 ViT-L/14. Each heatmap is independently normalized using its own minimum and maximum values. Yellow indicates higher values, while blue indicates lower values.}
    \label{fig:vitlnorm}
\end{figure}

We visualize these properties for DINOv2 ViT-L/14 in \Cref{fig:vitlatt} and \Cref{fig:vitlnorm} for attention and patch norm, respectively. We observe that both surgery and fine-tuning substantially reduce rogue token behavior compared with simple masking and the unmodified baseline, and fine-tuning is even slightly better. Although some rogue tokens remain visible in the attention
maps after surgery, foreground regions exhibit relatively
stronger [CLS]-to-patch attention and patch norm.

\subsection{DINOv2 ViT-g/14}
\begin{figure}[!ht]
    \centering
    \includegraphics[width=\columnwidth]{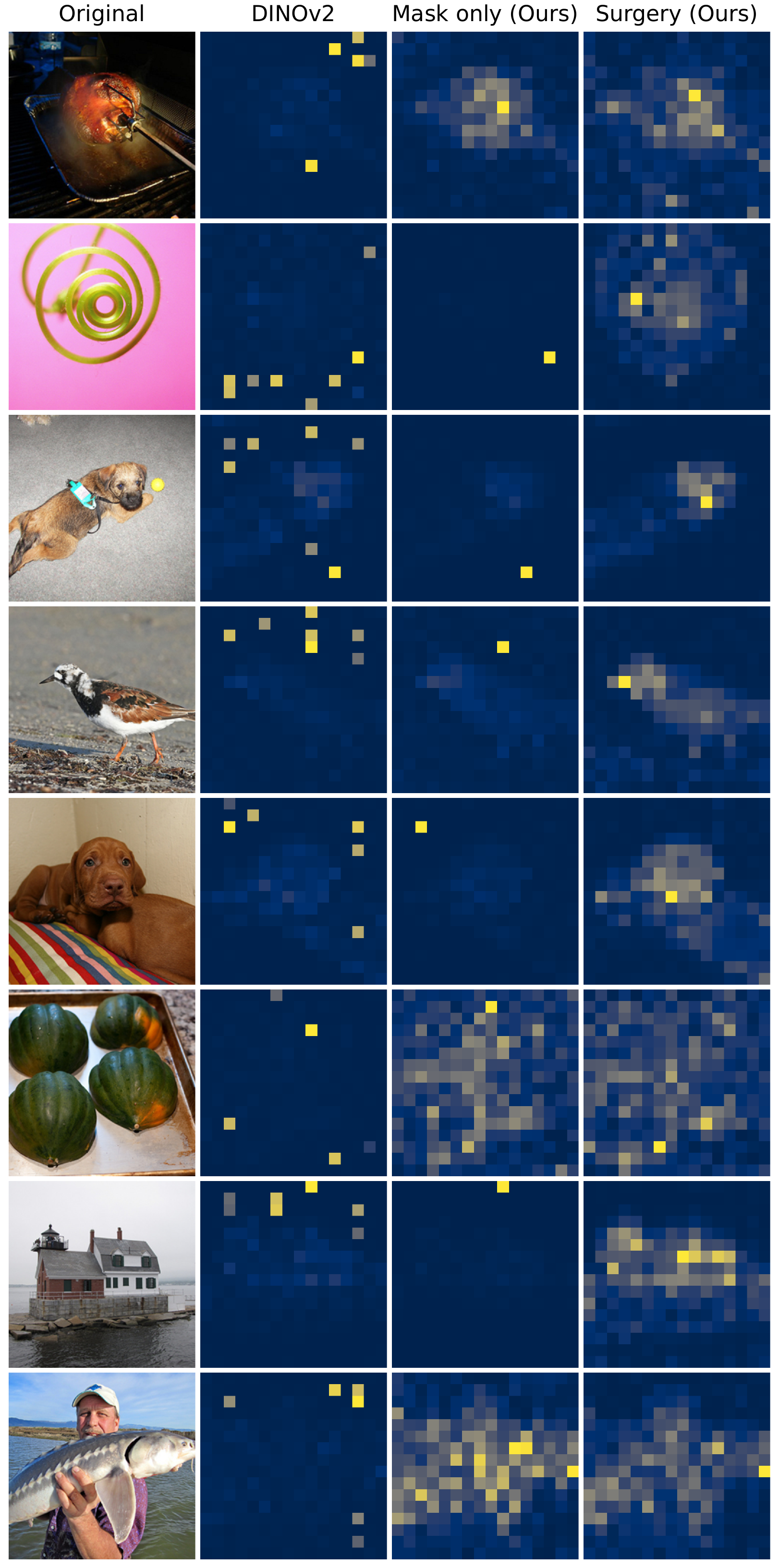}
    \caption{Comparison of [CLS]-to-patch attention across two proposed clean-up methods for DINOv2 ViT-g/14. Each heatmap is independently normalized using its own minimum and maximum values. Yellow indicates higher values, while blue indicates lower values.}
    \label{fig:vitgattention}
\end{figure}

\begin{figure}[!ht]
    \centering
    \includegraphics[width=\columnwidth]{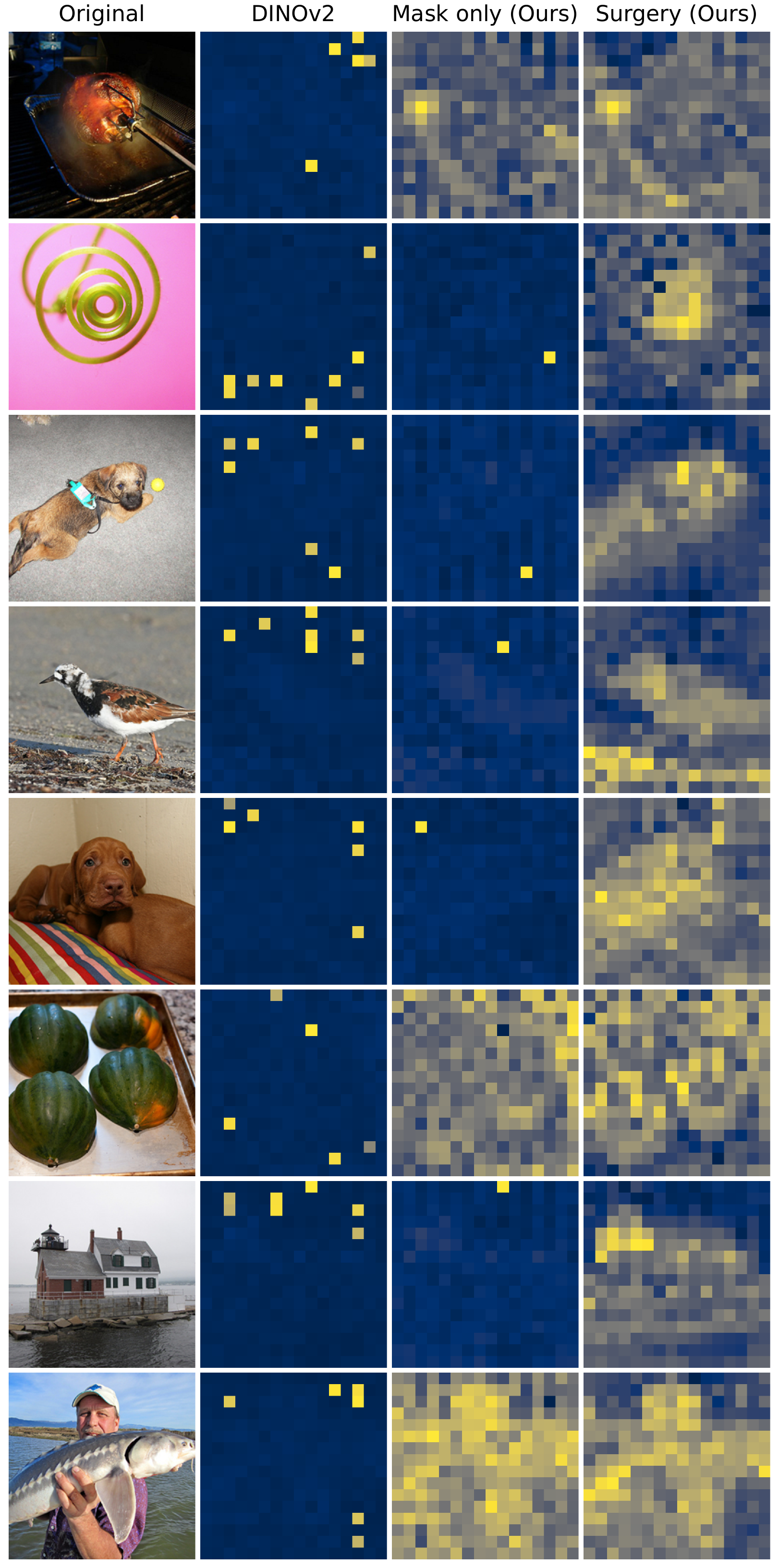}
    \caption{Comparison of patch norm across two proposed clean-up methods for DINOv2 ViT-g/14. Each heatmap is independently normalized using its own minimum and maximum values. Yellow indicates higher values, while blue indicates lower values.}
    \label{fig:vitgnorm}
\end{figure}

We further visualize the [CLS]-to-patch attention maps in \Cref{fig:vitgattention} and patch norms in \Cref{fig:vitgnorm}. Compared with DINOv2 ViT-L/14, DINOv2 ViT-g/14 exhibits more severe rogue-token behavior, with substantially more outliers in both attention maps and patch norms. Interestingly, both simple masking and our surgery method become more effective in this setting. For certain images, simple masking can even eliminate rogue-token behavior in both attention maps and patch-token norms. Nevertheless, our surgery method achieves the most consistent improvements across images, preserving relatively stronger [CLS]-to-patch attention and higher patch-token norms in foreground regions while substantially suppressing rogue-token dominance.
\subsection{OpenCLIP ViT-L/14}
\begin{figure}[!ht]
    \centering
    \includegraphics[width=\columnwidth]{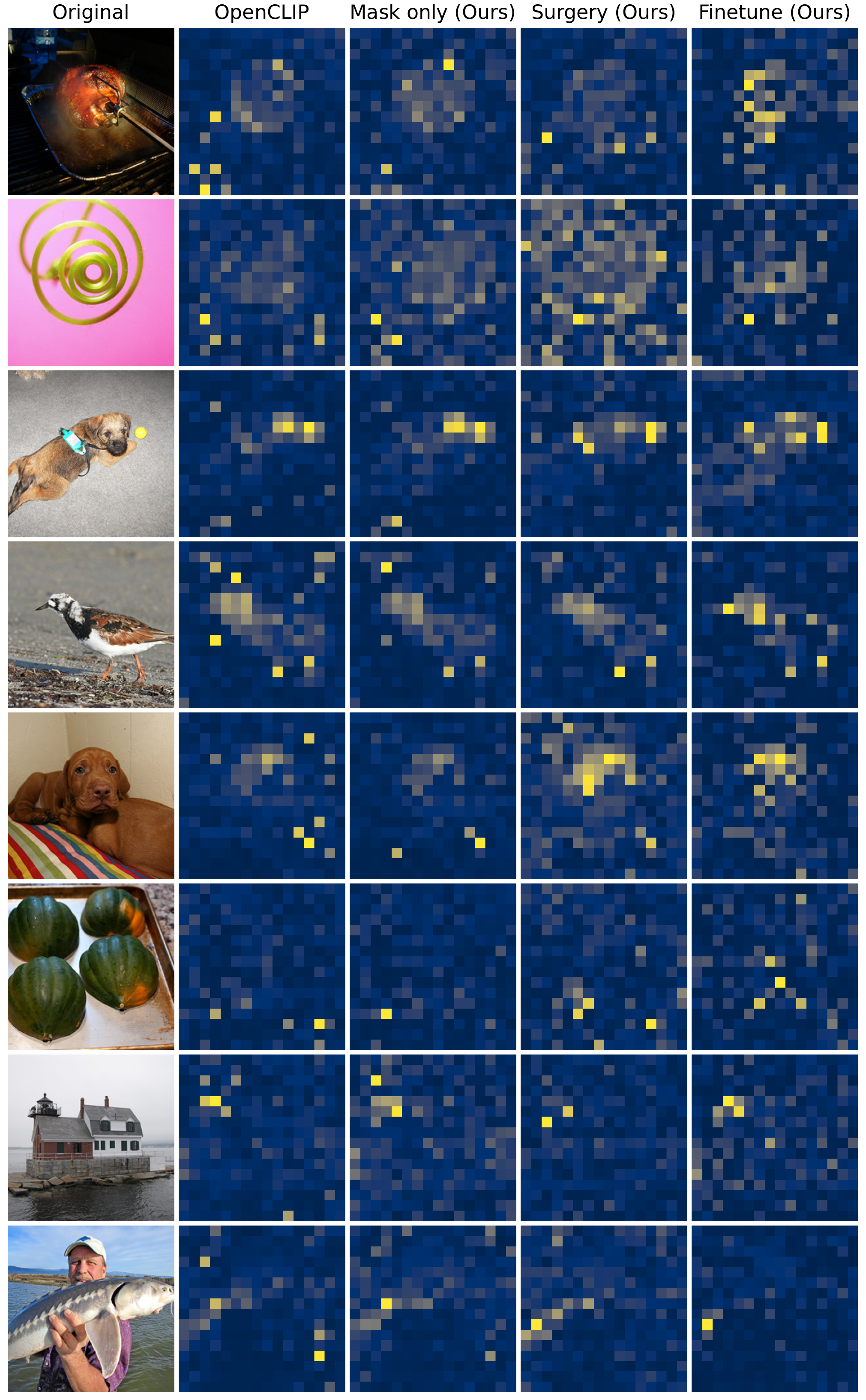}
    \caption{Comparison of [CLS]-to-patch attention across three proposed clean-up methods for OpenCLIP ViT-L/14. Each heatmap is independently normalized using its own minimum and maximum values. Yellow indicates higher values, while blue indicates lower values.}
    \label{fig:ocatt}
\end{figure}

\begin{figure}[!ht]
    \centering
    \includegraphics[width=\columnwidth]{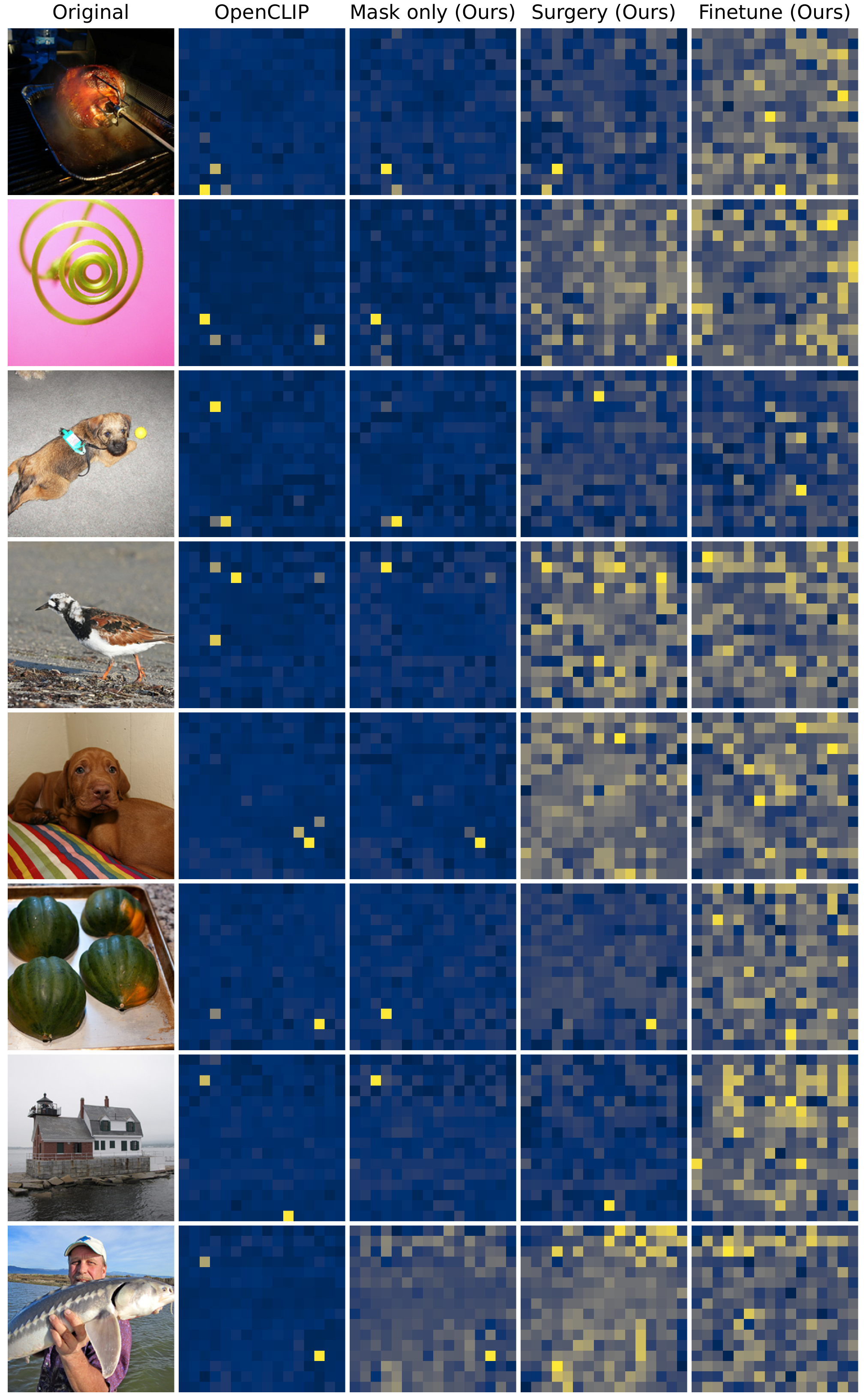}
    \caption{Comparison of patch norm across three proposed clean-up methods for OpenCLIP ViT-L/14. Each heatmap is independently normalized using its own minimum and maximum values. Yellow indicates higher values, while blue indicates lower values.}
    \label{fig:ocnorm}
\end{figure}

For OpenCLIP ViT-L/14, we visualize the [CLS]-to-patch attention maps in \Cref{fig:ocatt} and the patch norms in \Cref{fig:ocnorm}. Although the proposed surgery suppresses rogue-token dominance in the attention maps, some patches still exhibit relatively high norms, albeit with less pronounced dominance than in
the baseline and mask-only variants. This may explain the relatively small performance gap between the two methods on zero-shot segmentation. In contrast, fine-tuning not only suppresses the extreme patch-token norms but also restores more meaningful attention patterns, leading to improved segmentation performance. Notably, compared with DINOv2, patch norms appear to be less informative about foreground–background semantics and the effect of masking is less pronounced than in DINOv2, potentially due to differences in training strategies.

\clearpage
\newpage
\section{Visualization of Value-projection Columns}
\begin{figure}[p]
    \centering
    \includegraphics[width=\columnwidth]{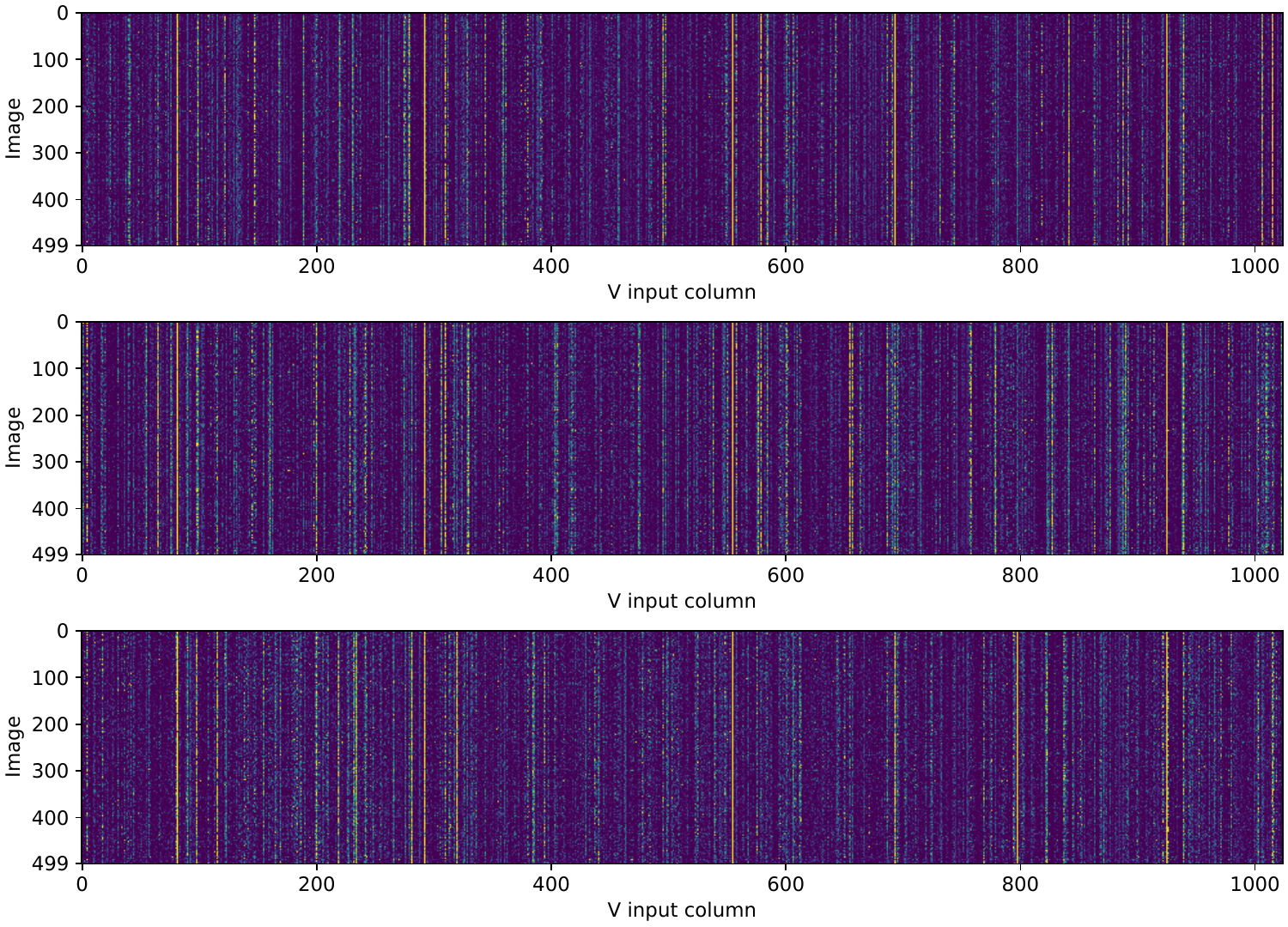}
    \caption{\textbf{CLS value-projection column contributions in OpenCLIP
    ViT-L/14.} We visualize layers 6, 7, and 8 from top to bottom
    for head index~0. Each row corresponds to one of the shared 500
    ImageNet-1k validation images at $224\times224$, in the same sampled
    order, each column indexes one of the 1024 columns of the value
    projection. Colors show normalized positive contributions before
    intervention. Brighter colors indicate larger normalized positive contributions.}
    \label{fig:cls_columns_openclip}
\end{figure}

\begin{figure}[p]
    \centering
    \includegraphics[width=\columnwidth]{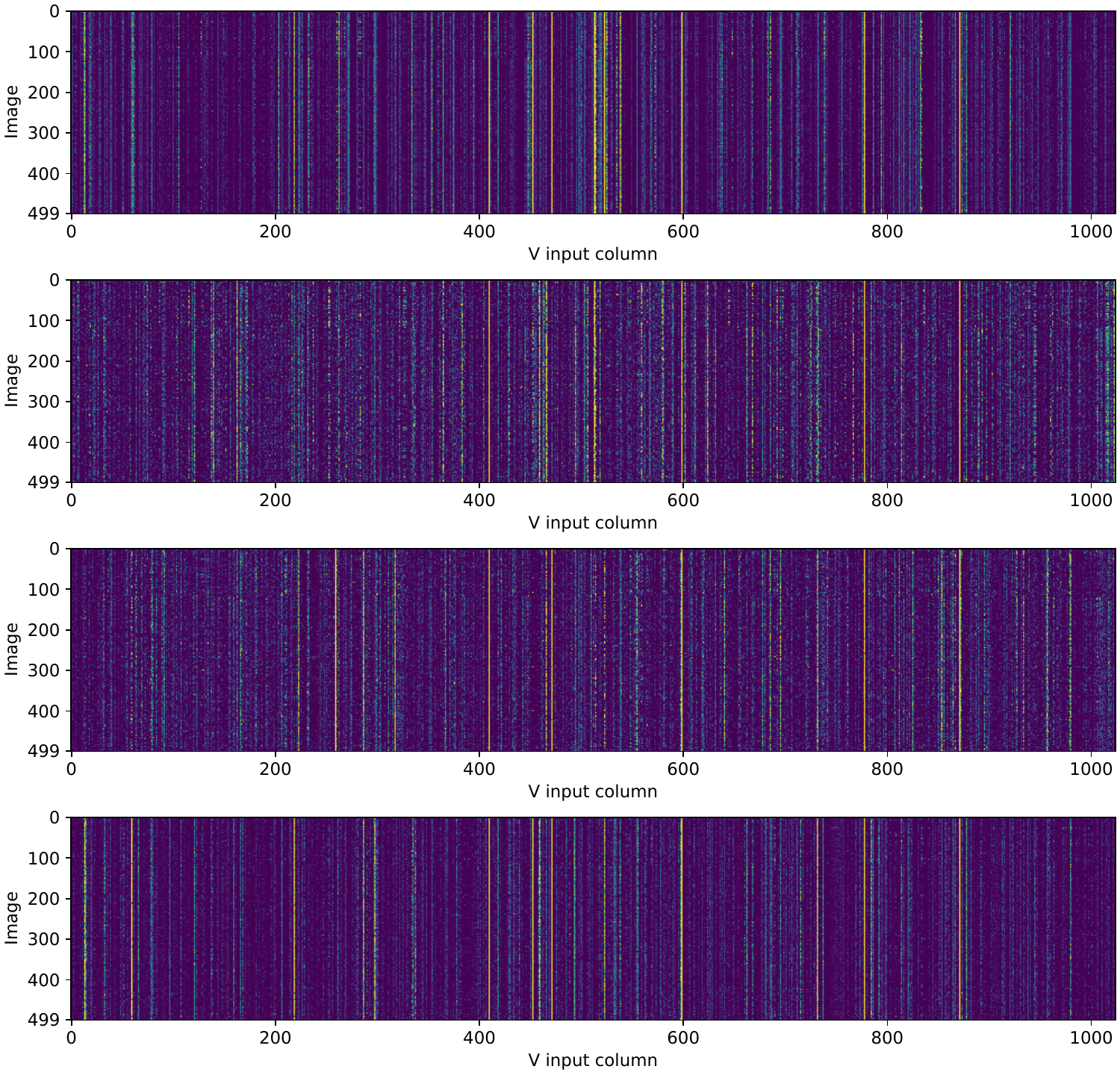}
    \caption{\textbf{CLS value-projection column contributions in DINOv2
    ViT-L/14.} We visualize layers 13, 14, 15, and 16 from top to bottom for head index~0. Each row corresponds to one of the same 500 ImageNet-1k
    validation images evaluated at $224\times224$, each column corresponds
    to one of the 1024 columns of the value projection. Colors show
    the per-image normalized positive contribution defined in
    Eq.~\eqref{eq:cls_positive_column_contribution}, measured before any
    intervention. Brighter colors indicate larger normalized positive contributions.}
    \label{fig:cls_columns_dinov2_l}
\end{figure}

\begin{figure}[p]
    \centering
    \includegraphics[width=\columnwidth]{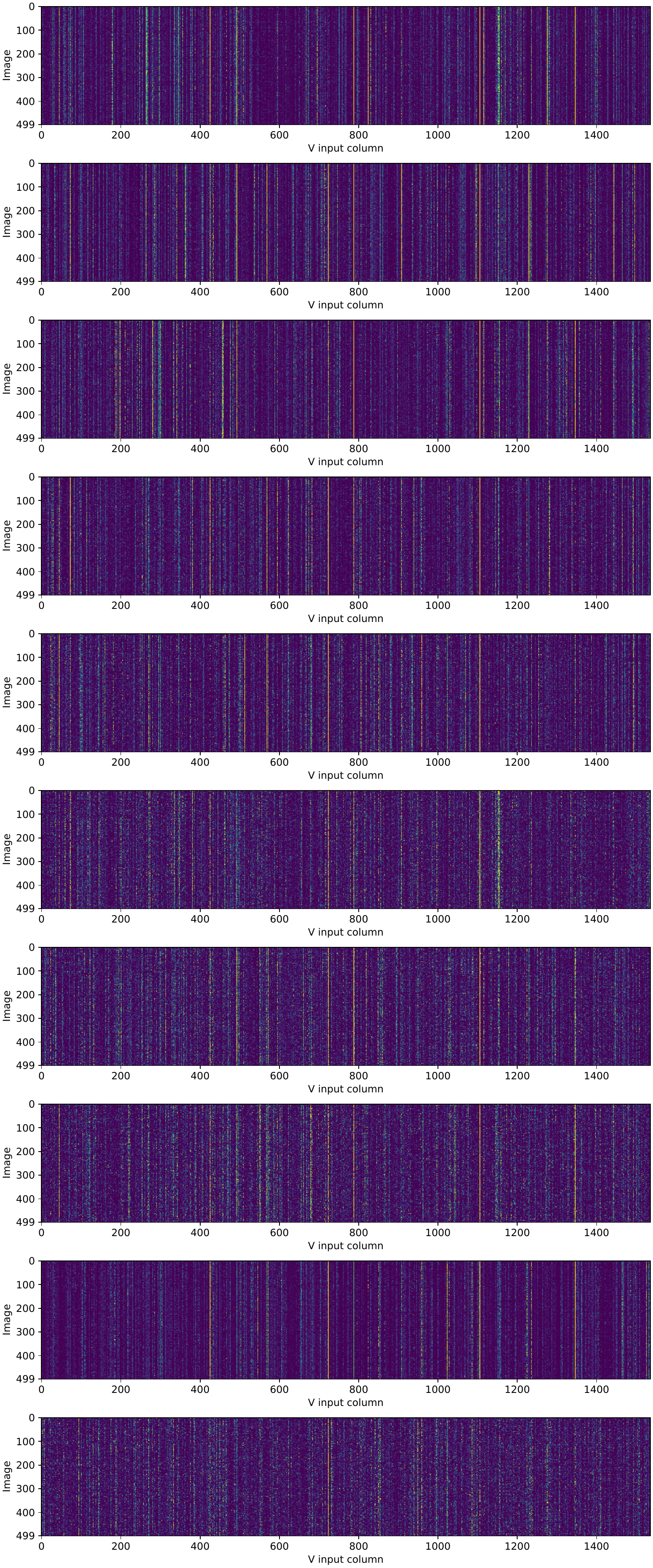}
    \caption{\textbf{CLS value-projection column contributions in DINOv2
    ViT-G/14.} We visualize layers 14--23 in ascending order from top to
    bottom for head index~0. Each row corresponds to one of the same 500 ImageNet-1k
    validation images evaluated at $518\times518$, each column corresponds
    to one of the 1536 columns of the value projection. Colors encode normalized positive contributions
    before intervention. Brighter colors indicate larger normalized positive contributions.}
    \label{fig:cls_columns_dinov2_g}
\end{figure}

In this section, we examine how the columns of the value-projection contribute to the value projection of the [CLS] token across different images. \Cref{fig:cls_columns_openclip}, \Cref{fig:cls_columns_dinov2_l}, and
\Cref{fig:cls_columns_dinov2_g} present this
diagnostic for OpenCLIP ViT-L/14, DINOv2 ViT-L/14, and DINOv2 ViT-G/14 respectively. All three models are evaluated on the same 500 images sampled from the ImageNet-1k validation set, retaining
the same sampled order. We use the pretrained backbones before applying Surgery or patch-to-[CLS] attention masking. The visualized layer ranges are 13--16 for DINOv2 ViT-L/14, 14--23 for DINOv2 ViT-G/14, and 6--8 for OpenCLIP ViT-L/14.

\paragraph{Contribution score.}
Let $\mathbf{x}_{n}^{(\ell)}\in\mathbb{R}^{D}$ be the [CLS] representation
after the first LayerNorm of layer~$\ell$ for image~$I_n$.
Let $\mathbf{W}_{V}^{(\ell,h)}\in\mathbb{R}^{d_h\times D}$ denote the
value-projection matrix of head~$h$, and let
$\mathbf{w}_{j}^{(\ell,h)}$ be its $j$-th input column. Excluding the
projection bias, we define
\begin{equation}
\begin{aligned}
    \mathbf{e}_{n}^{(\ell,h)}
    &= \mathbf{W}_{V}^{(\ell,h)}\mathbf{x}_{n}^{(\ell)},
    \\
    r_{n,j}^{(\ell,h)}
    &= x_{n,j}^{(\ell)}
       \left\langle \mathbf{e}_{n}^{(\ell,h)},
       \mathbf{w}_{j}^{(\ell,h)}\right\rangle,
    \\
    s_{n,j}^{(\ell,h)}
    &= \left[r_{n,j}^{(\ell,h)}\right]_{+},
    \\
    q_{n,j}^{(\ell,h)}
    &= \frac{s_{n,j}^{(\ell,h)}}
    {\max\!\left\{\sum_{k=1}^{D}
    s_{n,k}^{(\ell,h)},\,\varepsilon\right\}},
\end{aligned}
\label{eq:cls_positive_column_contribution}
\end{equation}
where $[a]_{+}=\max(a,0)$ and $\varepsilon=10^{-8}$, and the signed contributions satisfy
$\sum_{j=1}^{D}r_{n,j}^{(\ell,h)}
=\|\mathbf{e}_{n}^{(\ell,h)}\|_2^2$.
The displayed score $q_{n,j}^{(\ell,h)}$ retains their positive parts and normalizes them independently for each image. These maps therefore summarize contributions for each column weighted by each entry of the [CLS] token.

In each figure, the horizontal axis indexes input columns and the vertical axis indexes the 500 images. A bright vertical band indicates that the same input column repeatedly receives a large positive contribution across images.

\paragraph{OpenCLIP ViT-L/14.}
Figure~\ref{fig:cls_columns_openclip} illustrates that a small subset of columns consistently makes disproportionately large positive contributions to the [CLS] token representation across different images and layers. In contrast, the remaining columns exhibit weaker and more image-dependent contributions. Interestingly, the alignment of these dominant columns across layers further indicates that their positions remain relatively stable across layers, despite variations in the surrounding contribution patterns.

\paragraph{DINOv2 ViT-L/14.}
A similar pattern can be observed in DINOv2 ViT-L/14, as illustrated in Figure~\ref{fig:cls_columns_dinov2_l}. Narrow, bright vertical bands consistently appear at similar positions across images, while most surrounding columns exhibit relatively weak contributions. This further suggests that the concentration of strong positive contributions in a small subset of channels is not unique to OpenCLIP but also emerges in models trained with different learning objectives.

\paragraph{DINOv2 ViT-G/14.}
Figure~\ref{fig:cls_columns_dinov2_g} also shows clear vertical banding throughout the stack of layers. Prominent bands also remain visible at stable locations across images and can be traced across the layers. The weaker contributions surrounding these bands vary between layers. The resulting structure combines persistent column preferences with layer-dependent variation in how contributions are distributed over the remaining channels.

Taken together, these figures reveal consistent patterns of column dominance in [CLS] value projections across different pretrained backbones. The consistency of prominent vertical bands and the variation of weaker responses characterize how contributions are organized across images and layers. These observations provide a qualitative diagnostic for examining candidate columns, with the effects of intervention assessed separately through downstream evaluation.

\newpage
\clearpage

\end{document}